\documentclass[10pt]{article}
\usepackage{latexsym,amsmath,amssymb,graphics,amscd}
\usepackage{multirow}
\usepackage{soul}
\usepackage{booktabs}
\usepackage{tabularx}
\usepackage[hang,footnotesize]{caption}
\usepackage[pdftex]{graphicx}
\usepackage{graphicx}
\usepackage{overpic}
\usepackage{color}
\usepackage{cancel}
\usepackage[pdftex,breaklinks,colorlinks,
  citecolor=blue,
  urlcolor=blue]{hyperref}
\usepackage{tikz}
\usetikzlibrary{decorations.pathreplacing}
\usepackage{xr}
\addtocounter{footnote}{1}
\makeatletter
\@addtoreset{table}{bsection}
\def\thetable{\thesection.\@arabic\c@table}
\def\fps@table{h, t}
\@addtoreset{equation}{section}

\makeatother

\newtheorem{theorem}{Theorem}[section]

\newtheorem{proposition}[theorem]{Proposition}

\newsavebox{\savepar}

\makeatletter

\usepackage{scalerel,stackengine}
\stackMath
\newcommand\reallywidehat[1]{%
\savestack{\tmpbox}{\stretchto{%
  \scaleto{%
    \scalerel*[\widthof{\ensuremath{#1}}]{\kern-.6pt\bigwedge\kern-.6pt}%
    {\rule[-\textheight/2]{1ex}{\textheight}}
  }{\textheight}%
}{0.5ex}}%
\stackon[1pt]{#1}{\tmpbox}%
}

\begin{document}

\title{\textbf{Nonlinear memory capacity of parallel time-delay reservoir computers in the processing of multidimensional signals}}
\author{Lyudmila Grigoryeva$^{1}$, Julie Henriques$^{2}$, Laurent Larger$^{3}$, and Juan-Pablo Ortega$^{4, \ast}$}
\date{}
\maketitle

\begin{abstract}
This paper addresses the reservoir design problem in the context of delay-based reservoir computers  for multidimensional input signals, parallel architectures, and real-time multitasking. First, an approximating reservoir model is presented in those frameworks that provides an explicit functional link between the reservoir parameters and architecture and its performance in the execution of a specific task. Second, the inference properties of the  ridge regression estimator in the multivariate context is used to assess the impact of finite sample training on the decrease of the reservoir capacity.  Finally, an empirical study is conducted that shows the adequacy of the theoretical results with the empirical performances exhibited by various reservoir architectures in the execution of several nonlinear tasks with multidimensional inputs. Our results confirm the robustness properties of the parallel reservoir architecture with respect to task misspecification and parameter  choice that had already been documented in the literature.
\end{abstract}

\bigskip


\makeatletter
\addtocounter{footnote}{1} \footnotetext{%
Department of Mathematics and Statistics. Universit\"at Konstanz. Box 146. D-78457 Konstanz. Germany. {\texttt{Lyudmila.Grigoryeva@uni-konstanz.de} }}
\makeatother
\makeatletter
\addtocounter{footnote}{1} \footnotetext{%
CHRU Besan\c{c}on. 2 place Saint Jacques. F-25000 Besan\c{c}on. {\texttt{jhenriques@chu-besancon.fr} }}
\makeatother
\makeatletter
\addtocounter{footnote}{1} \footnotetext{%
FEMTO-ST, UMR CNRS  6174, Optics Department, Universit\'{e} de Franche-Comt\'{e}, UFR des
Sciences et Techniques. 15, Avenue des Montboucons. F-25000 Besan\c{c}on cedex. France. {\texttt{Laurent.Larger@univ-fcomte.fr} }}
\makeatother

\makeatletter
\addtocounter{footnote}{1} \footnotetext{%
Corresponding author. Centre National de la Recherche Scientifique, Laboratoire de Math\'{e}matiques de Besan\c{c}on, UMR CNRS 6623, Universit\'{e} de Franche-Comt\'{e}, UFR des
Sciences et Techniques. 16, route de Gray. F-25030 Besan\c{c}on cedex.
France. {\texttt{Juan-Pablo.Ortega@univ-fcomte.fr} }}
\makeatother

\medskip

\medskip

\medskip

\section{Introduction}
The recent and fast development of numerous massive data acquisition technologies results in a considerable growth of the data volumes that are stored and that need to be processed in the context of many human activities. The variability, complexity, and volume of this information have motivated the appearance of the generic term {\it Big Data}, which is mainly used to refer to  datasets whose features make the  traditional data processing approaches  inadequate. This relatively new  concept calls for the development of specialized tools for data preprocessing, analysis, transferring, and visualization, as well as for novel data mining and machine learning techniques in order to tackle  specific computational tasks. 
 
In this context, there is a recent but already well established paradigm for neural computation known by the name of {\bf{reservoir computing (RC)}}~\cite{jaeger2001, Jaeger04, maass1, maass2, Crook2007, verstraeten, lukosevicius} (also referred to as {\it Echo State Networks} and {\it Liquid State Machines}), that has already shown a significant potential in successfully confronting some of the challenges that we just described.

This brain-inspired machine learning methodology exhibits several competitive advantages with respect to more traditional approaches. First, the supervised learning scheme associated to it is extremely simple. Second, some implementations of the RC paradigm are based on the computational capacities of certain dynamical systems~\cite{Crutchfield2010} that open the door to physical realizations that have already been built using dedicated hardware (see, for instance,~\cite{jaeger2, Atiya2000, Appeltant2011, Rodan2011, Larger2012, Paquot2012}) and that, recently, have shown unprecedented information processing speeds~\cite{photonicReservoir2013}. Our work takes place in the context of this specific type of RCs and, more explicitly, in the so called {\bf time-delay reservoirs (TDRs)} that use the sampling of the solutions of time-delay differential equations in the construction of the RC. 

Despite the outstanding empirical performances of TDRs described in the above listed references and the convenience of the learning scheme associated to them, a well-known important drawback is that these devices show a certain lack of structural  task universality. More specifically, each task presented to a TDR requires that the TDR parameters and, more generally, its architecture are tuned in order to achieve optimal performance or, equivalently, small deviations from the optimal parameter values can seriously degrade the reservoir performance.   The optimal parameters have been traditionally found by trial and error or by running costly numerical scannings for each task. More recently, in~\cite{GHLO2014_capacity} we introduced a method to overcome this difficulty by providing a functional link between the RC parameters and its performance with respect to a given task and that can be used to accurately determine the optimal reservoir architecture by solving a well structured optimization problem; this feature simplifies enormously the implementation effort and sheds new light on the mechanisms that govern this information processing technique. 

This paper builds on the techniques introduced in~\cite{GHLO2014_capacity} and extends those results in the following directions:
\begin{description}
\item [(i)] The memory capacity formulas in~\cite{GHLO2014_capacity} are generalized to {\bf multidimensional input signals} and we provide capacity estimations for the simultaneous execution of several memory tasks. This feature, sometimes referred to as {\bf real-time multitasking}~\cite{maass2} is usually presented as one of the most prominent computational advantages of RC.
\item [(ii)] We provide memory capacity estimations for {\bf parallel arrays of reservoir computers}. This reservoir architecture has been introduced in~\cite{pesquera2012, GHLO2012} and has been empirically shown to exhibit improved robustness properties with respect to the dependence of the optimal reservoir parameters on the task presented to the device and also with respect to task misspecification.
\item [(iii)] We carry out an in-depth study of the ridge regression estimator in the multivariate context in order to assess {\bf  the impact of finite sample training on the decrease of reservoir capacity}. More specifically, when the teaching signal used to train the RC has finite size, the faulty estimation of the RC readout layer (see Section~\ref{The general setup for time-delay reservoir (TDR) computing}) introduces an error that is cumulated with the characteristic error associated to the RC scheme and that we explicitly quantify.
\item [(iv)] We conduct an empirical study that shows the adequacy of our theoretical results with the empirical performances exhibited by TDRs in the execution of various nonlinear tasks with multidimensional inputs. Additionally, we confirmed using the approximating model, the robustness properties of the parallel reservoir architecture with respect to task misspecification and parameter  choice that had already been documented in~\cite{GHLO2012}.
\end{description}

The paper is organized as follows: Section~\ref{Notation and preliminaries} recalls the general setup for time-delay reservoir computing, as well as the notions of characteristic error and memory capacity in the multitasking setup.
Section~\ref{The reservoir model for multidimensional computing and parallel architectures} constitutes the core of the paper and addresses the points {\bf (i)} through {\bf (iii)} listed in the previous paragraphs. The empirical study described in point {\bf (iv)} is contained in Section~\ref{Empirical study}.

\medskip

\noindent {\bf Acknowledgments:} We acknowledge partial financial support of the R\'egion de Franche-Comt\'e (Convention 2013C-5493), the European project PHOCUS (FP7 Grant
No. 240763), the ANR ``BIPHOPROC" project (ANR-14-OHRI-0002-02), and Deployment S.L. LG acknowledges financial support from the Faculty for the Future Program of the Schlumberger Foundation.

\section{Notation and preliminaries}
\label{Notation and preliminaries}

In this section we introduce the notation that we use in the paper, we briefly recall  the general setup for time-delay reservoirs (TDRs), and provide various preliminary concepts that are needed in the following sections.
\subsection{Notation}
Column vectors are denoted by bold lower or upper case  symbol like $\mathbf{v}$ or $\mathbf{V}$. We write $\mathbf{v} ^\top $ to indicate the transpose of $\mathbf{v} $. Given a vector $\mathbf{v} \in \mathbb{R}  ^n $, we denote its entries by $v_i$, with $i \in \left\{ 1, \dots, n
\right\} $; we also write $\mathbf{v}=(v _i)_{i \in \left\{ 1, \dots, n\right\} }$. 
The symbols $\mathbf{i} _n$ and $ \mathbf{0} _n $ stand for the vectors of length $n$ consisting of ones and zeros, respectively.  We denote by $\mathbb{M}_{n ,  m }$ the space of real $n\times m$ matrices with $m, n \in \mathbb{N} $. When $n=m$, we use the symbol $\mathbb{M}  _n $  to refer to the space of square matrices of order 
$n$. Given a matrix $A \in \mathbb{M}  _{n , m} $, we denote its components by $A _{ij} $ and we write $A=(A_{ij})$, with $i \in \left\{ 1, \dots, n\right\} $, $j \in \left\{ 1, \dots m\right\} $.   We write $\mathbb{I} _n $ and $\mathbb{O} _n $ to denote the identity matrix and the zero matrix of dimension $n$, respectively.
We use $\mathbb{S}_n   $ to  indicate the subspace $\mathbb{S}  _n \subset \mathbb{M}  _n $ of symmetric matrices, that is, $\mathbb{S}  _n = \left\{ A \in \mathbb{M}  _n \mid A ^\top = A\right\}$.
Given a matrix $ A \in \mathbb{M}  _{n , m} $, we denote by vec the operator that transforms $A$ into a vector of length $nm$ by stacking all its columns, namely, $${\rm vec}: \mathbb{M} _{n , m} 
\longrightarrow \mathbb{R}^{nm}, \quad {\rm vec} \left( A\right) = \left( A _{11}, \dots, A _{n1},  \dots, A_{1m}, \dots, A_{nm} \right) ^\top, \quad A \in \mathbb{M}  _{n , m}.$$ 
When $A$ is symmetric,
we denote by vech the operator that stacks    the elements on and below the main diagonal of $A$ into a vector of length $N := \frac{1}{2} n \left( n +1 \right) $, that is,
$${\rm vech}: \mathbb{S}  _n \longrightarrow   \mathbb{R}  ^N, \quad {\rm vech} \left( A\right) =\left( A_{11}, \dots, A_{n1}, A_{22}, \dots, A_{n2}, \dots, A_{nn}\right)  ^\top, \quad A \in \mathbb{S}  _n.$$ 
Let $N := \frac{1}{2}n \left( n+1 \right) $.  We denote by $L _n \in \mathbb{M}  _{N , n ^2 }$ and by $D _n \in \mathbb{M}  _{n ^2 , N}$ the 
elimination and the duplication matrices~\cite{luetkepohl:book}, respectively. These matrices satisfy that:
\begin{equation}
\label{Ln and Dn}
{\rm vech} \left( A\right) = L _n {\rm vec}\left(A\right),\quad \mbox{and} \quad
{\rm vec} \left( A \right) = D _n {\rm vech}\left( A \right).
 \end{equation}
Consider $A\in \mathbb{S}  _{n} $, $ \mathbf{v} = {{\rm vech}} \left( A\right)  \in \mathbb{R}  ^{N}$, and $S = \left\{\left( i, j\right) | i, j \in \left\{ 1, \dots, n\right\}, i \ge j  \right\}$. Let $\sigma : S \longrightarrow  \left\{ 1, \dots , N\right\}$ be the operator that assigns to the position of the entry $\left( i, j\right)$, $i \ge j$, of the matrix $A$ the position of the corresponding element of $\mathbf{v} $ in the  ${\rm vech}$ representation.  
We refer to the inverse of this operator as ${ {\sigma} ^{-1}  }:  \left\{ 1, \dots , N \right\}  \longrightarrow S $. The symbol $||A||_{\rm Frob}$ denotes the Frobenius norm of $A \in \mathbb{M} _{m,n}$ defined  as $\|A\|_{{\rm Frob}}^2:={\rm trace} \left(A^T A\right) $~\cite{Meyer:book:matrix}.  Finally, the symbols ${\rm E}[\cdot]$, ${\rm var}(\cdot)$, and ${\rm Cov}(\cdot,\cdot)$ denote the mathematical expectation, the variance, and the covariance, respectively.

\subsection{The general setup for time-delay reservoir (TDR) computing}
\label{The general setup for time-delay reservoir (TDR) computing}
{\bf The functional time-delay differential equations used for TDR computing. }The time-delay reservoirs studied in this paper are constructed by sampling the solutions of time-delay differential equations of the form
\begin{equation}
\label{time delay equation}
\dot{x} (t)= - x (t)+f(x (t- \tau), I(t), \boldsymbol{\theta}),
\end{equation}
where $f  $ is a nonlinear smooth function that will be referred to as  {\bf nonlinear kernel}, $\boldsymbol{\theta} \in \mathbb{R} ^K $ is a  vector that contains the parameters of the nonlinear kernel, $\tau >0$ is the {\bf delay}, $x (t) \in \mathbb{R} $, and $I (t)  \in \mathbb{R}$ is an external forcing that makes~\eqref{time delay equation} non-autonomous and that in our construction will be used as an inlet into the system for the signal that needs to be processed.
We emphasize that the solution space of equation \eqref{time delay equation} is infinite dimensional since an entire function $x \in C^{1}([- \tau, 0], \mathbb{R}) $ needs to be specified in order to initialize it. The nonlinear kernel $f$ is chosen based on the concrete physical implementation of the computing system that is envisioned. We consider two specific parametric sets of kernels that have already been explored in the literature, namely:
\begin{description}
\item [(i)]  The {\bf Mackey-Glass}~\cite{mackey-glass:paper} nonlinear kernel:
\begin{equation}
\label{Mackey-Glass nonlinear kernel}
f(x,I, \boldsymbol{\theta})= \frac{\eta \left(x+\gamma I \right)}{1+ \left(x+\gamma I\right)^p}, \enspace \boldsymbol{\theta}:=(\gamma, \eta, p) \in \mathbb{R} ^3,
\end{equation} 
which is used in electronics-based  RC implementations~\cite{Appeltant2011}.
\item  [(ii)] The {\bf Ikeda}~\cite{Ikeda1979} nonlinear kernel
\begin{equation}
\label{Ikeda nonlinear kernel}
f(x,I, \boldsymbol{\theta})= \eta \sin ^2 \left(x+ \gamma I+ \phi\right), \enspace \boldsymbol{\theta}:=(\eta, \gamma, \phi) \in \mathbb{R} ^3,
\end{equation}
associated to the optical RC implementations~\cite{Larger2012}.
\end{description}
For these specific choices of nonlinear kernel, the parameters $\gamma$ and $\eta $ are usually referred to as the {\bf input} and  {\bf feedback gains}, respectively. 

\medskip

\noindent{\bf Continuous and discrete-time approaches to multidimensional TDR computing. } We briefly recall the design of a TDR using the solutions of~\eqref{time delay equation}. The following constructions are discussed in detail in \cite{GHLO2014_capacity}. TDRs are based on the sampling of the solutions of~\eqref{time delay equation} when driven by an input forcing obtained out of the signal that needs to be processed. More specifically, let  ${\bf z}  (t) \in \mathbb{R} ^n $, $t \in \Bbb Z $,  be an $n$-dimensional discrete-time {\bf input signal}. This signal is, first, time and dimensionally multiplexed over a delay period by using an {\bf input mask} $C\in  \mathbb{M}_{N,n} $ and by setting $\mathbf{I}(t):=C {\bf z} (t)$, $t \in \Bbb Z $, where $N$ is a design parameter called the {\bf number of neurons} of each {\bf reservoir layer}. The resulting discrete-time $N$-dimensional signal $\mathbf{I}(t) \in  \mathbb{R}^N$ is called  {\bf input forcing}. 

We now consider two different constructions of the TDR depending on the way in which the solutions of the time-delay differential equation~\eqref{time delay equation} are handled. The {\bf  continuous time TDR} is constructed as a collection of {\bf neuron values} $x _i (t) $ organized in {\bf layers} $\mathbf{x} (t)  \in \mathbb{R}^N$ of $N\in \mathbb{N}$ of {\bf  virtual neurons} each, parameterized by $t \in \Bbb Z $. The value $x _i (t) $, referred to as the {\bf  $i$th neuron value} of the $t $th layer $\mathbf{x} (t) $ of the reservoir, is obtained by sampling a solution $x (t) $ of~\eqref{time delay equation} by setting
\begin{equation}
x_i (t) := x (t \tau- (N-i)d),\quad i \in \{1, \ldots, N\}, \quad t\in \Bbb Z,
\end{equation}
where $d:= \tau /N $ is referred to as the {\bf  separation between neurons}.
The solution $x (t)$ has been obtained by using an external forcing $I (s)$ in \eqref{time delay equation} constructed out of the input forcing $\mathbf{I} (t)  $ as follows: given $s \in \mathbb{R} $, let $t \in \Bbb Z  $  and $i \in \{2, \ldots, N\} $ be the unique values such that $s \in (t \tau- (N-i-1)d, t \tau- (N-i)d] $ and that  we use to define the external forcing as $I(s):=(\mathbf{I} (t) )_i$.

The {\bf discrete-time} TDR  is constructed via the Euler time-discretization of~\eqref{time delay equation} with an integration step of $d:= \tau/N $. In this case, the neuron values are determined by the following recursions:
\begin{equation}
\label{recursion euler}
x _i(t):= e^{-\xi}x_{i-1}(t)+(1-e^{-\xi})f(x _i(t-1), ({\bf I} (t)) _i, \boldsymbol{\theta}),
\end{equation}
with $x _0(t):= x _N(t-1)$, $ \xi:=\log (1+ d)$, and $i\in \left\{ 1, \dots, N \right\}$. In this case, the recursions~\eqref{recursion euler} uniquely determine a smooth map
$F: \mathbb{R}^N\times \mathbb{R}^N \times \mathbb{R} ^K \rightarrow \mathbb{R}^N $ referred to as the {\bf reservoir map}
that specifies the neuron values of a given $t$th layer as a recursion on the neuron values of the preceding layer $t-1$ via an expression of the form 
\begin{equation}
\label{vector discretized reservoir main}
\mathbf{x}(t) = F( \mathbf{x} (t-1), \mathbf{I} (t), \boldsymbol{ \theta }). 
\end{equation}

\medskip

\begin{figure}[h!]
\hspace{-0.5cm}
\includegraphics[scale=.27, angle=0]{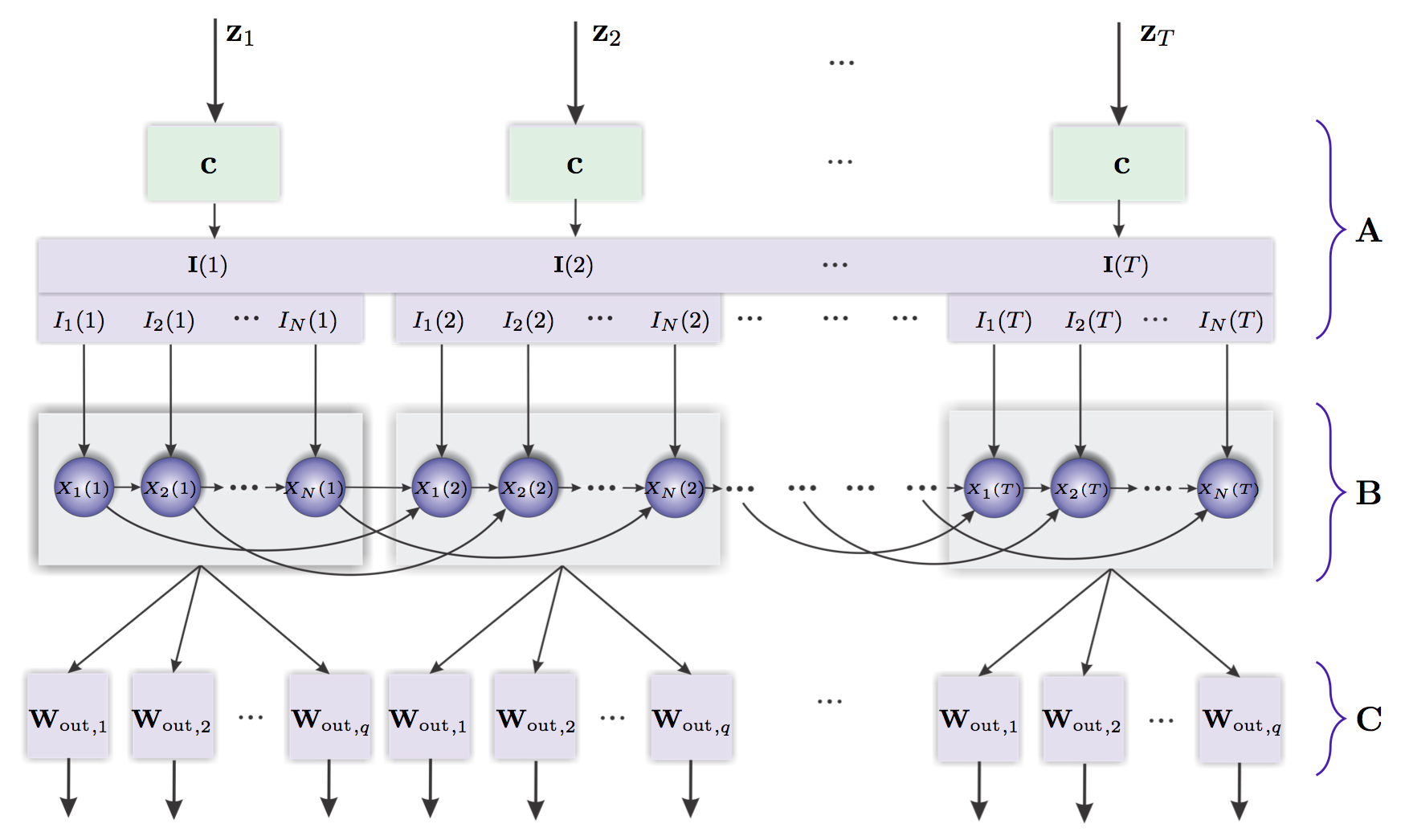}
\caption{Diagram representing the architecture of a TDR reservoir with a multitask readout. The module A is the input layer, B is a neural diagram representing the  discrete-time  reservoir dynamics implied by equation~\eqref{recursion euler}, and C is the multitask readout layer in which each column of the matrix $W_{{\rm out}} $ accomplishes a different task based on the same reservoir output.}
\label{RC figure multitasking}
\end{figure}

\noindent{\bf The TDR memory capacity for real-time multitasking. } In this paper we study the performance of TDRs at the time of simultaneously performing several memory tasks (see Figure~\ref{RC figure multitasking}). This means that we will evaluate the ability of the TDR to reproduce a prescribed multidimensional nonlinear function of the input signal
\begin{align}
\label{memory task def}
\begin{array}{cccc}
H: &\mathbb{R} ^{(h+1)n}&\longrightarrow &\mathbb{R}^q\\
&{\rm vec}( {\bf z}  (t) , \dots,  {\bf z}  (t - h)  )&\longmapsto& \mathbf{y} (t),
\end{array}
\end{align}
that we will call {\bf $q$-dimensional $h$-lag memory  task} for the $n$-dimensional input signal $\left\{  {\bf z}  (t)\right\}_{t \in \Bbb Z} $. 

In the RC context, this task is performed by using a finite size realization of the input signal $\{ {\bf z}(-h+1), \ldots, {\bf z} (T) \}   $ that is used to construct a $q$-dimensional {\bf teaching signal} $\{ {\bf y}(1), \ldots, {\bf y} (T) \}   $ by setting $ {\bf y}(t):=H\left({\rm vec}( {\bf z}  (t) , \dots,  {\bf z}  (t - h)  ) \right) $. The teaching signal is subsequently used to determine a pair $\left(  { W}_{{\rm out}}, \boldsymbol{a}_{\rm out} \right) \in \mathbb{M} _{N,q} \times  \mathbb{R} ^q  $  that performs the memory task as an affine combination of the reservoir outputs.  The optimal pair $\left(  { W}_{{\rm out}}, \boldsymbol{a}_{\rm out} \right)$ is obtained with a ridge regression that minimizes  the regularized square error, that is,
 \begin{align}
({ W}_{{\rm out}}, \boldsymbol{a}_{\rm out})&=\mathop{\rm arg\, min}_{{ W}   \in \mathbb{M}_{N,q} ,  \boldsymbol{a}\in\mathbb{R} ^q  } \left({\rm trace} \left({\rm E} \left[  ({ W} ^{ \top} \cdot {\bf x}(t)  +\boldsymbol{a} -\mathbf{y} (t)  )^\top ({ W} ^{ \top} \cdot {\bf x}(t)  +\boldsymbol{a} -\mathbf{y} (t)  )\right] \right)  + \lambda \|{ W}   \|^2_{\rm Frob}\right)\\
&=:\mathop{\rm arg\, min}_{{ W}   \in \mathbb{M}_{N,q} ,  \boldsymbol{a}\in\mathbb{R} ^q  } \left({\rm MSE}(W, \boldsymbol{a}) + \lambda \|{ W}   \|^2_{\rm Frob}\right).
\label{RC optimization problem statement}
\end{align}
The optimal pair $({ W}_{{\rm out}}, \boldsymbol{a}_{\rm out}) $ that solves the ridge regression problem~\eqref{RC optimization problem statement} is referred to as the {\bf readout layer}.
The ridge regularization parameter $\lambda \in \mathbb{R} $ is usually tuned during the training phase via cross-validation. The explicit solution of the optimization problem \eqref{RC optimization problem statement} (see \cite{GHLO2014_capacity} for the details) is given by 
\begin{align}
\label{linear_cov_system_gamma}
 { W}_{\rm out} = & (\Gamma (0) + \lambda \mathbb{I}  _N ) ^{-1}{\rm Cov} (\mathbf{x} (t), \mathbf{y} (t) ),\\
\label{linear_cov_system_c}
\boldsymbol{a}_{\rm out} = & \boldsymbol{\mu} _y - { W}_{\rm out} ^\top \boldsymbol{\mu} _x,  
\end{align}
where $\boldsymbol{\mu} _x:={\rm E}[\mathbf{x} (t)] \in \mathbb{R} ^N $, $\boldsymbol{\mu} _y :={\rm E}[\mathbf{y} (t)] \in \mathbb{R} ^q $, $\Gamma (0) :={\rm Cov} (\mathbf{x} (t), \mathbf{x} (t) ) \in \Bbb S _N$, and ${\rm Cov}(\mathbf{x} (t), \mathbf{y} (t) ) \in \mathbb{M}_{N,q}$. Stationarity hypotheses are assumed on the teaching signal and the reservoir output so that the first and second order moments that we just listed are time-independent.
The error committed by the reservoir when accomplishing the task $H$ with the optimal readout will be referred to as its {\bf characteristic error} and is given by the expression 
\begin{align}
\label{error rc optimal}
{\rm MSE}(W_{{\rm out}}, \boldsymbol{a}_{{\rm out}})={\rm trace}({\rm Cov} \left(\mathbf{y} (t), \mathbf{y} (t)\right) -{ W}_{{\rm out}}^\top (\Gamma (0) + 2\lambda \mathbb{I} _N) { W}_{{\rm out}}),
\end{align}
that can be encoded under the form of a {\bf memory capacity} $C_H( \boldsymbol{\theta}, C, \lambda)$ with values between zero and one that depends on the task $H$ that is being tackled, the input mask $C $, the reservoir parameters $\boldsymbol{\theta} $ and the regularization constant $\lambda $:
\begin{align}
\label{capacity formula}
C_H( \boldsymbol{\theta}, C, \lambda):=&1- \dfrac{{\rm MSE}(W_{{\rm out}}, \boldsymbol{a}_{{\rm out}})}{{\rm trace} \left({\rm Cov}(\mathbf{y} (t), \mathbf{y} (t) )\right)}=\dfrac{{\rm trace}( { W}_{{\rm out}}^\top (\Gamma (0) + 2\lambda \mathbb{I} _N) { W}_{{\rm out}})}{{\rm trace} \left({\rm Cov}(\mathbf{y} (t), \mathbf{y} (t) )\right)},
\end{align}
where $W_{\rm out}$ is provided by the solution in \eqref{linear_cov_system_gamma}. 
In order to evaluate~\eqref{capacity formula} for a specific memory task, the expressions of  $\Gamma (0)  $, ${\rm Cov} (\mathbf{x} (t) , \mathbf{y} (t)) $, and ${\rm Cov}(\mathbf{y} (t), \mathbf{y} (t) )$  need to be computed. 
The matrix $\Gamma (0)  $ depends exclusively on the input signal and the reservoir architecture but ${\rm Cov}(\mathbf{y} (t), \mathbf{y} (t) )$ and ${\rm Cov} (\mathbf{x} (t) , \mathbf{y} (t)) $ are related to the specific memory task $H$ at hand. The computation of~\eqref{capacity formula} is in general very complicated
and that is why in
\cite{GHLO2014_capacity} we introduced a simplified reservoir model that allowed us to efficiently evaluate it for one-dimensional statistically independent input signals and memory tasks. The extension of this theoretical tool to a multidimensional setup and to parallel architectures is one of the main goals of this paper. 

Finally, there are situations in which the moments $\boldsymbol{\mu} _x$, $\boldsymbol{\mu} _y$, $\Gamma (0)$, and ${\rm Cov}(\mathbf{x} (t), \mathbf{y} (t) )$, necessary to compute the readout layer $({ W}_{{\rm out}}, \boldsymbol{a}_{\rm out}) $ using the equations~\eqref{linear_cov_system_gamma}-\eqref{linear_cov_system_c}, are obtained directly out of finite sample realizations of the teaching signal and of the reservoir output. The use in that context of finite sample empirical estimators carries in its wake an additional error that adds up to the characteristic error~\eqref{error rc optimal} and that we study later on in Section~\ref{Finite sample effect in the Tikhonov regularised reservoir regression}.

\section{Reservoir memory capacities for parallel architectures and for multidimensional input signals and tasks}
\label{The reservoir model for multidimensional computing and parallel architectures} 
In this section we generalize the reservoir model  proposed in \cite{GHLO2014_capacity} in order to accommodate the treatment of multidimensional input signals and the computation of memory capacities associated to the execution of several simultaneous memory tasks, as well as the performance evaluation of parallel reservoir architectures. The main virtue of the reservoir model is that it allows for the explicit computation of the different elements that constitute the capacity formula~\eqref{capacity formula} making hence accessible its evaluation.

In the last section we study the dependence of the reservoir performance on the length of the teaching signal  and the regularization regression parameter. In particular we formulate asymptotic expressions that  provide an estimation of the added error that is commited in memory tasks when the training is incomplete due to the finiteness of the training sample.

The reservoir model introduced in \cite{GHLO2014_capacity} is based on the observation that optimal reservoir performance is frequently attained when the reservoir is functioning in a neighborhood of an asymptotically stable equilibrium of the autonomous system associated to~\eqref{time delay equation}. This feature suggests the possibility of approximating the reservoir by 
its partial linearization at that stable fixed point with respect to the delayed self feedback but keeping the nonlinearity at the level of the input signal injection. This observation motivated 
in \cite{GHLO2014_capacity} an in-depth study of the stability properties of the equilibria $x_0$ of the time-delay differential equation \eqref{time delay equation} and of the corresponding fixed points $\mathbf{x} _0 = x _0 \mathbf{i} _N \in \mathbb{R} ^N $ of the discrete-time approximation \eqref{vector discretized reservoir main}, both considered in the autonomous regime, that is, when $I(t)=0$ in \eqref{time delay equation} and $\mathbf{I} (t)= {\bf 0} _N $ in \eqref{vector discretized reservoir main}, respectively. In particular, it was shown that (see Corollary D.5 and Theorem D.10 in the Supplementary Material in \cite{GHLO2014_capacity}) that
$|\partial _{{x}}f({ x} _0, { 0} , \boldsymbol{\theta}  )| < 1$ is a sufficient condition for the asymptotic stability of $x_0\in \mathbb{R}$ and $\mathbf{x} _0 = x _0 \mathbf{i} _N \in \mathbb{R} ^N $ in the continuous and discrete-time cases, respectively, which given a particular kernel $f$ allows for the identification of specific regions in parameter space in which stability is guaranteed (see Corollaries~D.6 and D.7 of the Supplementary Material  in \cite{GHLO2014_capacity} for the Mackey-Glass and  Ikeda kernel cases).

\subsection{The reservoir model for multidimensional input signals}
\label{The approximated model of TDR with multidimensional input signals} 

Consider a discrete-time TDR described by a reservoir map $F: \mathbb{R}^N\times \mathbb{R}^N \times \mathbb{R} ^K \rightarrow \mathbb{R}^N $ as in~\eqref{vector discretized reservoir main}. Let  $\mathbf{x} _0 \in \mathbb{R} ^N $ be a stable fixed point  of the autonomous systems associated to~\eqref{vector discretized reservoir main}, that is,    $F( {\bf x} _0  , {\bf 0} _N , \boldsymbol{ \theta }) = {\bf x} _0 $. In order to write down the approximate reservoir model as in \cite{GHLO2014_capacity} we start by approximating \eqref{vector discretized reservoir main} by its partial linearization at $\mathbf{x} _0 $ with respect to the delayed self feedback and by the $R$th-order Taylor series expansion on the input forcing $\mathbf{I}  (t) \in \mathbb{R} ^N $. We obtain the following expression: 
\begin{equation}
\label{linearization of the reservoir}
\mathbf{x} (t) = F(\mathbf{x} _0, {\bf 0}_N, \boldsymbol{ \theta } ) + A (\mathbf{x} _0, \boldsymbol{ \theta }) (\mathbf{x} (t-1) - \mathbf{x} _0)  + \boldsymbol{\varepsilon}  (t),
\end{equation}
where  $F(\mathbf{x} _0, {\bf 0}_N, \boldsymbol{ \theta } ) $ is the reservoir map evaluated at the point $(\mathbf{x} _0,{\bf 0} _N, \boldsymbol{ \theta })$ and $A (\mathbf{x} _0, \boldsymbol{ \theta }) :=D_{\mathbf{x} } F(\mathbf{x} _0,{\bf 0} _N, \boldsymbol{ \theta }) $ is the first derivative of $F$ with respect to its first argument, computed at the point $(\mathbf{x} _0,{\bf 0} _N, \boldsymbol{ \theta })$.
The vector $\boldsymbol{\varepsilon} (t) \in \mathbb{R} ^N $, $t\in \Bbb Z $, in \eqref{linearization of the reservoir} is obtained out of the Taylor series expansion of $F(\mathbf{x} (t), \mathbf{I} (t), \boldsymbol{\theta} )$ in \eqref{vector discretized reservoir main} on $\mathbf{I} (t)$ up to some fixed order $R\in \mathbb{N} $. For each $r \in \left\{ 1, \dots, N \right\} $ its $r$th component can be written as
\begin{equation}
\label{eps_t_r}
{ \varepsilon } _r (t) = (1 - { e} ^{- \xi})\sum^{R}_{i = 1} \dfrac{1 }{i !}  (\partial _{{I} } ^{(i)} f   )({x} _0 , 0, \boldsymbol{\theta}  ) \sum^{r}_{j = 1} e^{-(r-j) \xi  }  I_j (t)^i,
\end{equation}
where $(\partial _{{I} } ^{(i)} f   )({x} _0 , 0, \boldsymbol{\theta}  )$ is the $i$th order partial derivative of the nonlinear reservoir kernel map $f$ in \eqref{time delay equation} with respect to the second argument $I(t)$ computed at the point $({x} _0 , 0, \boldsymbol{\theta}  )$. Finally, $A (\mathbf{x} _0, \boldsymbol{ \theta }) $
is  called the {\bf  connectivity matrix} of the reservoir at the point $\mathbf{x} _0 $ and has the following explicit form
\begin{align}
\label{DxF}
&A(\mathbf{x} _0, \boldsymbol{ \theta })=  
{\small\left(
\begin{array}{cccccc}
 \Phi&    0 &\hdots &0&  { e^{- \xi }}{}\\
 { e^{- \xi }}\Phi&   \Phi&  \hdots &0& { e^{-2 \xi }}{}\\
  e^{- 2\xi }\Phi&  e^{- \xi }\Phi&  \hdots &0&  { e^{- 3\xi }}{}\\
  \vdots&\vdots&\ddots&\vdots&\vdots\\
  e^{- (N-1)\xi }\Phi&   e^{- (N-2)\xi }\Phi&   \hdots &{ e^{- \xi }}{} \Phi &\Phi + { e^{- N\xi }}{}
\end{array}
\right)},
\end{align}
where $ \Phi :=(1- e ^{- \xi })\partial _{x}f(x _0, 0 , \boldsymbol{\theta}  )$ and $\partial _{x}f(x _0, 0 , \boldsymbol{\theta}  ) $ is the first derivative of the nonlinear kernel $f$ in \eqref{time delay equation} with respect to the first argument and computed at the point $(x _0, 0 , \boldsymbol{\theta}  )$.  

Suppose now that the input signal  is a collection of $n$-dimensional independent and identically distributed random variables $\left\{ {\bf z}  (t)\right\}_{t \in \Bbb Z }\sim {\rm IID} ({\bf 0} _n , \Sigma _z )$, $\Sigma _z\in \Bbb S ^+ _n$, and that we take as input mask the matrix $C \in \mathbb{M} _{N,n}$. Since for each ${t \in \Bbb Z }$ the input  forcing $\mathbf{I} (t) \in \mathbb{R} ^N $ is constructed via the assignment $\mathbf{I}(t):=C {\bf z} (t)$ we have that $ \left\{ \mathbf{I} (t) \right\}_{t\in \Bbb Z} \sim {\rm IID} ({\bf 0} _N , \Sigma _I  )$, with $\Sigma _I :=C \Sigma _z C^\top$. It follows then that $I_j(t) = \sum^{n}_{k = 1} C_{jk} z_k (t) $ which substituted in \eqref{eps_t_r} yields that
\begin{equation}
\label{eps_t_V}
\boldsymbol{ \varepsilon } (t) = (1 - { e} ^{- \xi})\left(\begin{array}{c}V _R \left( {\bf z} (t), \left\{ C_{1,\cdot}\right\}, x _0 , \boldsymbol{\theta}  \right) \\V _R \left( {\bf z} (t),\left\{ C_{j,\cdot} \right\}_{j\in \left\{ 1,2\right\} },x _0 , \boldsymbol{\theta}\right)\\ \vdots\\ V _R \left( {\bf z} (t),\left\{ C_{j,\cdot} \right\}_{j\in \left\{ 1,\dots, N\right\} },x _0 , \boldsymbol{\theta}\right) \end{array}\right),
\end{equation}
with the polynomials 
\begin{align}
\label{V_R_polynomial}
V _R \left( {\bf z} (t), \left\{ C_{j,\cdot} \right\}_{j\in \left\{ 1,\dots, r\right\} },x _0 , \boldsymbol{\theta}\right) &:= \sum^{R}_{i = 1} (\partial _{{I} } ^{(i)} f   )({x} _0 , 0, \boldsymbol{\theta}  ) \sum^{r}_{j = 1} { e} ^{-(r-j)\xi} \sum^{}_{k _1 +\dots + k _n  = i} \dfrac{1}{k _1! \dots k _n!} \prod^{n}_{s=1} C_{js}^{k_s} \cdot \prod^{n}_{s=1} z_{s}(t)^{k_s}.
\end{align}
The symbol  $\left\{ C_{j,\cdot} \right\} $, $j\in \left\{ 1, \dots N\right\} $,  denotes the set of all the entries in the $j$th row of the input  mask matrix $C$. The assumption $\left\{ {\bf z}  (t)\right\}_{t \in \Bbb Z }\sim {\rm IID} ({\bf 0} _n , \Sigma _z )$ implies that $\{\boldsymbol{ \varepsilon } (t)\}_{t \in \Bbb Z} $ is also a family of $N$-dimensional independent and identically distributed random variables  with  mean $\boldsymbol{\mu}_{{ \varepsilon }} $ and covariance matrix $\Sigma _{{ \varepsilon }} $ given by
\begin{equation}
\label{mu_eps_res_main_r}
(\boldsymbol{\mu}_{{ \varepsilon }} )_r= (1 - { e} ^{- \xi})\sum^{R}_{i = 1}   (\partial _{{I} } ^{(i)} f   )({x} _0 , 0, \boldsymbol{\theta}  ) \sum^{r}_{j = 1} e^{-(r-j) \xi  }  \sum^{}_{k _1 +\dots + k _n  = i} \dfrac{1}{k _1! \dots k _n!} \prod^{n}_{s=1} C_{js}^{k_s} \cdot \mu _{k _1 , \dots, k _n }({\bf z} ),
\end{equation}
where 
\begin{equation}
\label{higher order moment}
\mu _{k _1 , \dots, k _n }({\bf z}(t)  ) :={\rm E} \left[\prod^{n}_{s=1} z _s (t) ^{k_s} \right]
\end{equation}
denotes a higher-order moment of ${\bf z} (t) \in \mathbb{R} ^n $ whose existence we assume for values $k _1 , \dots, k _n  $ such that $k _1 + \cdots+ k _n \leq 2R $.   Additionally, $\Sigma _{{ \varepsilon }}:={\rm E}\left[ (\boldsymbol{ \varepsilon } (t)  - \boldsymbol{\mu}_{{ \varepsilon }}) (\boldsymbol{ \varepsilon } (t)  - \boldsymbol{ \mu}_{{ \varepsilon }}) ^\top \right]$  has entries determined by the relation:
\begin{align}
\label{Sigma_eps_main}
(\Sigma _{{ \varepsilon }}) _{rs} = (1 - { e} ^{- \xi }) ^2  {\rm E}&\left[ V_R \left(  {\bf z} (t), \left\{ C_{j,\cdot}\right\}_{j \in \left\{ 1, \dots, r\right\} },x _0 , \boldsymbol{\theta} \right) \cdot V_R \left(  {\bf z} (t), \left\{ C_{j,\cdot}\right\}_{j \in \left\{ 1, \dots, s\right\} },x _0 , \boldsymbol{\theta} \right)  \right] \nonumber \\
&- (\boldsymbol{\mu} _{\boldsymbol{\varepsilon }})_r (\boldsymbol{\mu} _{\boldsymbol{\varepsilon }})_s  , \quad r,s \in \left\{ 1, \dots, N \right\},
\end{align}
where the first summand is computed by first multiplying the polynomials $V_R \left(  {\bf z}, \left\{ C_{j,\cdot}\right\}_{j \in \left\{ 1, \dots, r\right\} },x _0 , \boldsymbol{\theta} \right)$ and $ V_R \left(  {\bf z} , \left\{ C_{j,\cdot}\right\}_{j \in \left\{ 1, \dots, s\right\} },x _0 , \boldsymbol{\theta} \right) $  on the variable ${\bf z}\in \mathbb{R}^n $ and subsequently evaluating the resulting polynomial according to the following convention:  any monomial of the form $a z _1^{k _1} \cdots z _n^{k _n} $ is replaced by $a \mu _{k _1 , \dots, k _n }({\bf z}  ) $.

A particular case in which the higher-order moments~\eqref{higher order moment} can be readily computed is when  $\left\{ {\bf z}  (t)\right\}_{t \in \Bbb Z }\sim {\rm IN} ({\bf 0} _n , \Sigma _z )$,  that is, $\left\{ {\bf z}  (t)\right\}_{t \in \Bbb Z } $  follows an $n$-dimensional multivariate normal distribution. Indeed, following~\cite{Holmquist1988, Gaussian_moments2003}, let   $k _{1}, \ldots, k _{n} \in \mathbb{N}$ be $n$ nonzero natural numbers such that  $K:= k_{1}+k_{2}+\dots+k_{n}$ and let ${\bf z}^{\mathcal{K}} := \left(z _{1}  \mathbf{i} _{k_{1}}^\top , z _{2}  \mathbf{i} _{k_{2}}^\top , \dots, z _{n}  \mathbf{i} _{k_{n}}^\top\right)^\top \in \mathbb{R}^K $.  The vector ${\bf z}^{\mathcal{K}} \in \mathbb{R}^K $ is  Gaussian  with zero mean and covariance matrix $\Sigma _z ^{\mathcal{K}} \in \Bbb S _K $ given by $(\Sigma _z ^{\mathcal{K}})_{ij} = {\rm Cov}({ z}_i^{\mathcal{K}},{ z}_j^{\mathcal{K}} )$, for any $i,j \in \left\{ 1, \dots, K \right\} $, that is, ${\bf z}^{\mathcal{K}} \sim {\rm IN} ({\bf 0} _K, \Sigma _z ^{\mathcal{K}})$. 
Then, using Theorem 1 in \cite{Gaussian_moments2003}, we can write that
\begin{equation}
\label{mu_I_res}
\mu _{k _{1}, \ldots, k _{n} }({\bf z}  ) = \left\{ \begin{array}{cl}
 Hf( \Sigma _z ^{\mathcal{K}}   ),&  {\rm when} \enspace K= 2l,\quad l \in \mathbb{N},  \\
  & \\
  0,&{\rm otherwise},
\end{array}
\right.
\end{equation}
where   the symbol $Hf( \Sigma _z ^{\mathcal{K}})$ denotes the hafnian of the covariance matrix $ \Sigma _z ^{\mathcal{K}}$ of order $2l$, $l\in \mathbb{N} $, defined by
\begin{equation}
\label{hafnian}
Hf(\Sigma _z ^{\mathcal{K}}) := \sum^{}_{I,J}  (\Sigma _z ^{\mathcal{K}})_{i'_1 j'_1} (\Sigma _z ^{\mathcal{K}})_{i'_2 j'_2}\cdots (\Sigma _z ^{\mathcal{K}})_{i'_l j'_l},
\end{equation}
where the sum is running over all the possible decompositions of $\left\{ 1, 2, \dots, 2l=K\right\} $ into disjoint subsets $I, J $  of the form $I= \left\{i'_1,\ldots,i'_l \right\} $, $J= \left\{j'_1,\ldots,j'_l  \right\} $, such that $i'_1 <\cdots<i'_l$,  $j'_1 <\cdots<j'_l$, and $i'_w<j'_w$, for each $w\in \{1, \ldots, l\}$.

We now proceed as in \cite{GHLO2014_capacity} and consider~\eqref{linearization of the reservoir} as a VAR(1) model~\cite{luetkepohl:book} driven by the independent noise $\{\boldsymbol{ \varepsilon } (t)\}_{t \in \Bbb Z} $. If we assume that the nonlinear kernel $f$ satisfies the stability condition $|\partial _{{x}}f({ x} _0, { 0} , \boldsymbol{\theta}  )| < 1$, then the proof of Theorem D.10 in~\cite{GHLO2014_capacity}  shows that the spectral radius $\rho(A(\mathbf{x} _0,  \boldsymbol{\theta} )) <1$, which implies in turn that~\eqref{linearization of the reservoir} has a unique causal and second order stationary solution~\cite[Proposition 2.1]{luetkepohl:book} $\{ \mathbf{x} (t)\}_{t \in \Bbb Z } $ with time-independent mean 
\begin{equation}
\label{mu_x}
\boldsymbol{\mu} _{{x} } =  (I _N -  A(\mathbf{x} _0,  \boldsymbol{\theta} ) )^{-1} (F(\mathbf{x} _0 , {\bf 0} _N , \boldsymbol{\theta} ) - A(\mathbf{x} _0,  \boldsymbol{\theta} )  \mathbf{x} _0  + \boldsymbol{\mu} _{{\varepsilon}}).
\end{equation}
The model  \eqref{linearization of the reservoir} can hence be rewritten in mean-adjusted form as 
\begin{equation}
\label{VAR model reservoir}
\mathbf{x} (t) - \boldsymbol{\mu} _{{x} } = A(\mathbf{x} _0,  \boldsymbol{\theta} )( \mathbf{x} (t-1) - \boldsymbol{\mu}  _{{x} }) + (\boldsymbol{\varepsilon} (t) - \boldsymbol{\mu}_{{\varepsilon}}).
\end{equation}
Additionally, the autocovariance function $\Gamma (k):= {\rm E}\left[\left(\mathbf{x}(t)-\boldsymbol{\mu} _{{x} } \right) \left(\mathbf{x}(t-k)-\boldsymbol{\mu} _{{x} } \right)^\top\right]$ at lag $k \in \Bbb Z $ is determined  by the Yule-Walker equations~\cite{luetkepohl:book}, which have the following solutions in vectorized form: 
\begin{align}
\label{Gamma0 single reservoir} 
{\rm vech}(\Gamma(0) )&= \left(\mathbb{I}_{N'}-L_N (A(\mathbf{x} _0,  \boldsymbol{\theta} )\otimes A(\mathbf{x} _0,  \boldsymbol{\theta} )\right))^{-1} {\rm vech}(\Sigma _{{ \varepsilon }}),\\
\Gamma(k)&=A(\mathbf{x} _0,  \boldsymbol{\theta} ) \Gamma(k-1) \enspace {\rm with} \enspace \Gamma(-k)= \Gamma(k)^\top,
\end{align}
where $N' := \dfrac{1}{2}N (N+1)$ and $L_N \in \mathbb{M} _{N', N^2}$, $D_N \in   \mathbb{M} _{N^2, N'}$ are the elimination and the duplication matrices, respectively. We recall that the autocovariance function $\Gamma (0)$ is one of the key components of the capacity formula~\eqref{capacity formula} that we intend to explicitly evaluate.

\subsection{The reservoir model for parallel TDRs with multidimensional input signals}
\label{The approximating model of parallel TDRs with multidimensional input signals} 
\begin{figure}[h!]
\hspace{-0.5cm}
{\footnotesize
\begin{overpic}[scale = 0.37,tics = 5]{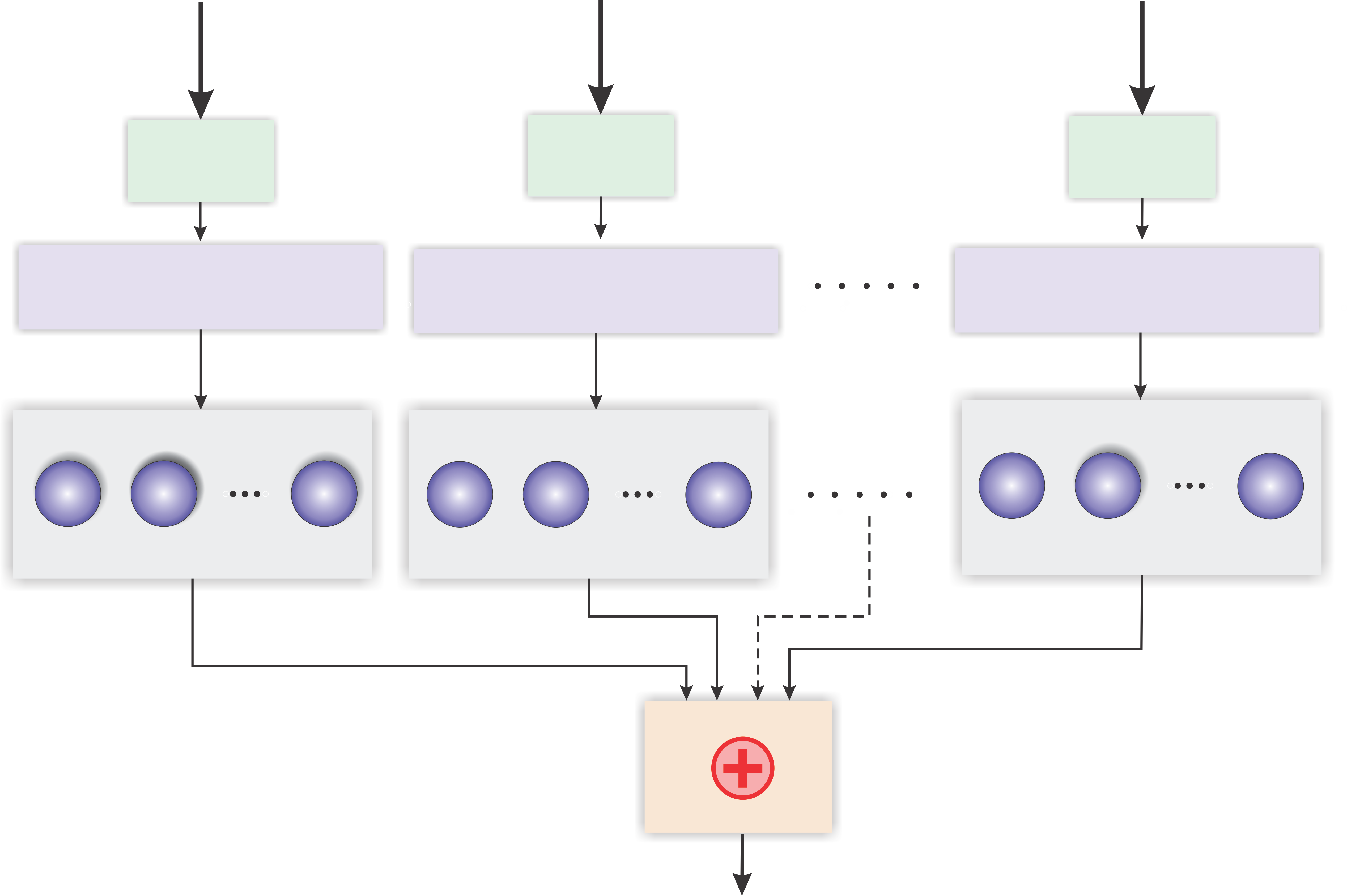}
\!\!
\put (12,54) {\large$C^1$}
\put (42,54) {\large$C^2$}
\put (82.5,54) {\large$C^p$}
\put (2.2,29.5) {\fontsize{5}{90}$X^1_{1}(T)$}
\put (9.5,29.5) {\fontsize{5}{100}$X^1_{2}(T)$}
\put (21,29.5) {\fontsize{5}{100}$X^1_{N_1}(T)$}

\put (31.3,29.5) {\fontsize{5.4}{100}$X^2_{1}(T)$}
\put (38.3,29.5) {\fontsize{5.4}{100}$X^2_{2}(T)$}
\put (50.3,29.5) {\fontsize{5.4}{100}$X^2_{N_2}(T)$}

\put (72.3,30) {\fontsize{5.4}{100}$X^p_{1}(T)$}
\put (79.3,30) {\fontsize{5.4}{100}$X^p_{2}(T)$}
\put (91.4,30) {\fontsize{1}{100}$X^p_{N_p}(T)$}

\put (15.3,39) {\scriptsize$R(\boldsymbol{\theta}_1, \lambda_1)$}
\put (44.7,39) {\scriptsize$R(\boldsymbol{\theta}_2, \lambda_2)$}
\put (85,39) {\scriptsize$R(\boldsymbol{\theta}_p, \lambda_p)$}

\put (16,65) {$\mathbf{Z}$}
\put (46,65) {$\mathbf{Z}$}
\put (86,65) {$\mathbf{Z}$}

\put (13.5,44.5) {\large$\mathbf{I}^1$}
\put (43,44.5) {\large$\mathbf{I}^2$}
\put (83.5,44.5) {\large$\mathbf{I}^p$}

\put (48,6) {\large$W_{\rm out}$}
\end{overpic} }
\caption{Diagram representing the architecture of a parallel reservoir computer.}
\label{parallel reservoir figure}
\end{figure}
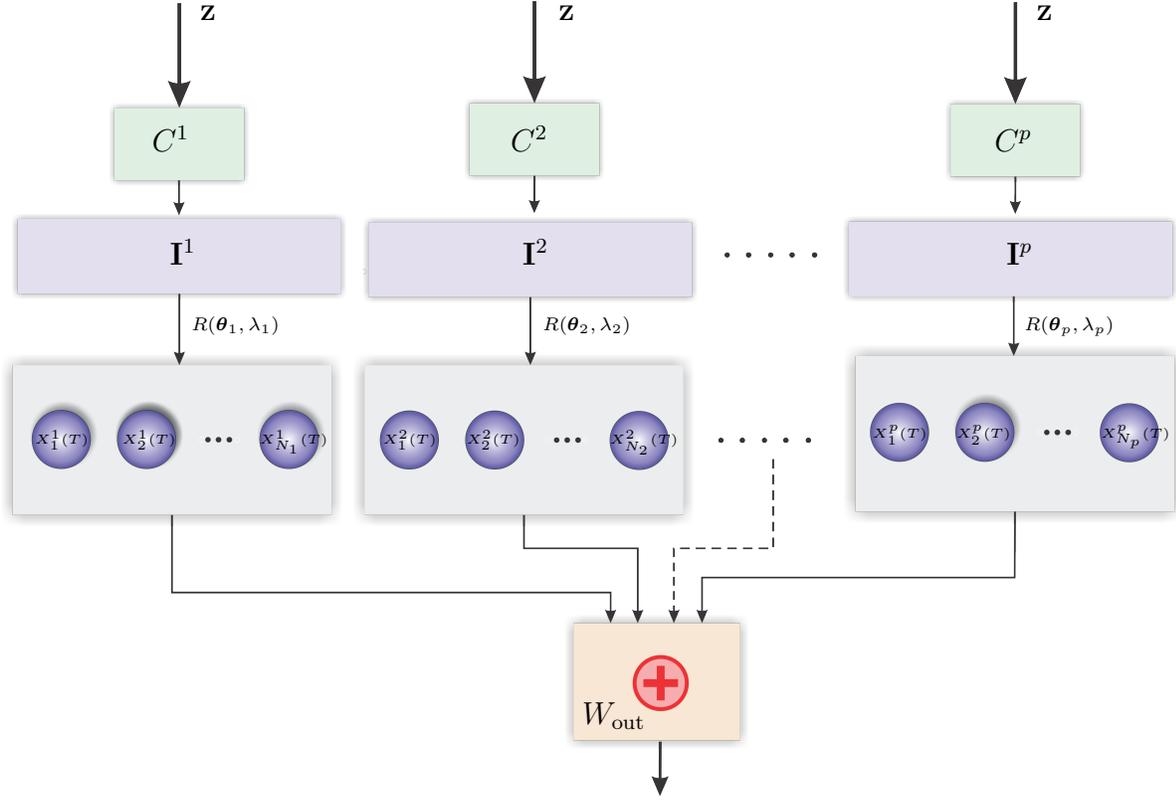

In this section we generalize the reservoir model that we just developed to the parallel time-delay reservoir architecture that introduced in~\cite{pesquera2012, GHLO2012}. This reservoir design  has shown very satisfactory robustness properties with respect to model misspecification and parameter choice. The basic idea on which this approach is built consists of presenting the input signal to a parallel array of reservoirs, each of them running with different parameter values. The concatenation of the outputs of these reservoirs is then used to construct  a single readout layer via a ridge regression.
Figure \ref{parallel reservoir figure} provides a diagram representing the parallel reservoir computing architecture.

\medskip

\noindent{\bf Discrete-time description  of the parallel TDRs. } Consider a parallel array of  $p$ time-delay reservoirs as in Figure \ref{parallel reservoir figure}. For each $j\in \left\{ 1, \dots, p \right\} $, the $j$th time-delay reservoir is based on a time-delay differential equation like \eqref{time delay equation}, has an associated nonlinear kernel $f^{(j)}$ that depends on the parameters vector $\boldsymbol{\theta} ^{(j)} \in \mathbb{R} ^{K_j}$ and the time-delay $\tau _j>0 $, namely
\begin{equation}
\label{time delay equation jth reservoir}
\dot{x} (t)= - x (t)+f^{(j)}(x (t- \tau_j), I(t), \boldsymbol{\theta}^{(j)}).
\end{equation}
Let  $N_j \in \mathbb{N} $ be the number of the virtual neurons of the $j$th reservoir and let $d_j:= \tau_j/N _j $ be the corresponding separation between neurons. Let $N ^\ast  := \sum^{p}_{j = 1} N_j$ and $K ^\ast  := \sum^{p}_{j = 1} K_j$ be the total number of virtual neurons and the total number of  parameters of the parallel array, respectively. The discrete-time description of the parallel array of $p$ TDRs with  total number of neurons $N ^\ast $ is obtained by Euler time-discretizing each of the differential equations~\eqref{time delay equation jth reservoir} with integration step $d_j$ and by organizing the solutions in neural layers described by the following recursions
\begin{equation}
\label{recursion euler parallel}
x _i^{(j)}(t):= e^{-\xi^{(j)}}x_{i-1}^{(j)}(t)+(1-e^{-\xi^{(j)}})f^{(j)}(x _i^{(j)}(t-1), ({\bf I}^{(j)} (t)) _i, \boldsymbol{\theta}^{(j)}), \quad i \in \{1, \ldots,N _j\},\  j \in \{1, \ldots,p\}
\end{equation}
with $\xi ^{(j)} := {\rm log}(1+d_{j}) $ and using the convention $x _0^{(j)}(t):= x ^{(j)}_{N_j}(t-1)$. The different $p$ input forcings ${\bf I}^{(j)}(t) \in \mathbb{R}^{N _j}$, $j \in \{1, \ldots,p\} $, are created out of the $n $-dimensional input signal $\{{\bf z} (t)\}_{t \in \Bbb Z}$ by using $p$ input masks $C ^{(j)} \in \mathbb{M}_{N _j, n}$ and by setting ${\bf I}^{(j)}(t):=C ^{(j)} {\bf z}(t)  $.

Consider now $\mathbf{X} (t) = \left(\mathbf{x} ^{(1)} (t), \dots, \mathbf{x} ^{(p)} (t) \right)  \in \mathbb{R} ^{N ^\ast  } $, where  $\mathbf{x} ^{(j)} (t) \in \mathbb{R} ^{N_j}$, $j \in \left\{ 1, \dots, p\right\} $, is the neuron layer at time $t$ corresponding to the $j$th individual TDR. 
As in the case of the individually operating time-delay reservoir in Section~\ref{The approximated model of TDR with multidimensional input signals},  
the recursions~\eqref{recursion euler parallel} uniquely determine  reservoir maps $F^{(j)}: \mathbb{R}^{N _j}  \times \mathbb{R}^{N_j}   \times \mathbb{R} ^{K_j}   \rightarrow \mathbb{R}^{N_j}   $, constructed out of the associated nonlinear kernels $f^{(j)}$, $j \in \left\{ 1, \dots, p \right\} $ that can be put together to determine the map
\begin{equation}
\label{vector discretized parallel reservoir main vector}
\left( 
\begin{array}{c} 
\mathbf{x} ^{(1)} (t) \\ 
\mathbf{x} ^{(2)} (t) \\ 
\vdots \\
\mathbf{x} ^{(p)} (t)
\end{array}\right) = \left( 
\begin{array}{c} 
F^{(1)} (\mathbf{x} ^{(1)} (t-1), \mathbf{I} ^{(1)}(t), \boldsymbol{\theta}  ^{(1)}) \\ 
F ^{(2)} (\mathbf{x} ^{(2)} (t-1), \mathbf{I}^{(2)} (t), \boldsymbol{\theta}  ^{(2)}) \\ 
\vdots \\
F^{(p)} (\mathbf{x} ^{(p)} (t-1), \mathbf{I} ^{(p)}(t), \boldsymbol{\theta}  ^{(p)})
\end{array}\right).
\end{equation}
that can be rewritten as 
\begin{equation}
\label{vector discretized parallel reservoir main}
\mathbf{X}(t) = F( \mathbf{X} (t-1), \mathbf{I} (t), \boldsymbol{ \Theta }) , 
\end{equation}
where $F: \mathbb{R}^{N ^\ast}  \times \mathbb{R}^{N^\ast}   \times \mathbb{R} ^{K ^\ast}   \rightarrow \mathbb{R}^{N^\ast}   $ is referred to as the {\bf parallel reservoir map}, $\mathbf{I} (t):= \left(\mathbf{I}^{(1)}(t), \ldots, \mathbf{I}^{(p)}(t)\right)$, and
 $\boldsymbol{ \Theta }:=(\boldsymbol{\theta}  ^{(1)}, \dots, \boldsymbol{\theta} ^{(p)} )\in \mathbb{R} ^{K ^\ast  } $ is the vector containing all the parameters of the parallel array of TDRs. Parallel TDRs based on the recursion~\eqref{vector discretized parallel reservoir main} are referred to in the sequel as {\bf discrete-time parallel TDRs}.
 
 We now generalize to the parallel context the reservoir model that we introduced in Section~\ref{The approximated model of TDR with multidimensional input signals}. We start by choosing $p$ stable equibria ${x} _0 ^{(j)}$ of the dynamical  systems~\eqref{time delay equation jth reservoir} or, equivalently, $p $ fixed points of the form ${\bf x} _0^{(j)} := x_0 ^{(j)}\mathbf{i} _{N_j} $ of each of the reservoir maps in~\eqref{vector discretized parallel reservoir main vector}. These fixed points determine a fixed point  ${\bf X} _0 := (  {\bf x} _0^{(1)}, \dots, {\bf x} _0^{(p)} ) \in \mathbb{R} ^{N ^\ast }$ of the parallel array in \eqref{vector discretized parallel reservoir main}. Now, as in Section~\ref{The approximated model of TDR with multidimensional input signals} we partially linearize \eqref{vector discretized parallel reservoir main} at ${\bf X} _0$ and use a higher $R$th-order Taylor series expansion on the forcing $\mathbf{I}  (t) \in \mathbb{R} ^{N ^\ast  } $. Analogously to the single reservoir case, we obtain that 
\begin{equation}
\label{linearization of the parallel reservoir}
\mathbf{X} (t) = F(\mathbf{X} _0, {\bf 0}_{N^\ast  }, \boldsymbol{ \Theta } ) + A (\mathbf{X} _0, \boldsymbol{ \Theta }) (\mathbf{X} (t-1) - \mathbf{X} _0)  + \boldsymbol{\varepsilon}^{({\bf X} _0, \boldsymbol{\Theta} )}  (t),
\end{equation}
where  $A({\bf X} _0, \boldsymbol{\Theta}  ) := D_{\mathbf{X} } F({\bf X} _0, {\bf 0}_{N^\ast  },   \boldsymbol{\Theta} )$ is the {\bf parallel reservoir connectivity matrix},  which is the first  derivative of $F$ with respect to its first argument and evaluated at the point $({\bf X} _0, \boldsymbol{\Theta})$, and
\begin{equation}
\label{parallel reservoir epsilon}
 \boldsymbol{ \varepsilon } (t) ^{({\bf X} _0, \boldsymbol{\Theta} )}:=\left( \begin{array}{c} \boldsymbol{\varepsilon} (t) ^{({\bf x} _0^{(1)}, \boldsymbol{\theta}^{(1)} )}\\\boldsymbol{\varepsilon} (t) ^{({\bf x} _0^{(2)}, \boldsymbol{\theta} ^{(2)})}\\\vdots\\\boldsymbol{\varepsilon} (t) ^{({\bf x} _0^{(p)}, \boldsymbol{\theta} ^{(p)})}\end{array} \right) \in \mathbb{R} ^{N ^\ast}
\end{equation}
with
\begin{equation}
\label{eps_t_V jth reservoir}
\boldsymbol{ \varepsilon } (t) ^{({\bf x} _0^{(j)}, \boldsymbol{\theta}^{(j)} )} := (1 - { e} ^{- \xi^{(j)}})\left(\begin{array}{c}V^{(j)} _R \left( {\bf z} (t), \left\{ C^{(j)}_{1,\cdot}\right\}, x _0^{(j)} , \boldsymbol{\theta}^{(j)}  \right) \\V ^{(j)}_R \left( {\bf z} (t),\left\{ C^{(j)}_{i,\cdot} \right\}_{i\in \left\{ 1,2\right\} },x _0^{(j)} , \boldsymbol{\theta}^{(j)}\right)\\ \vdots\\ V^{(j)} _R \left( {\bf z} (t),\left\{ C^{(j)}_{i,\cdot} \right\}_{i\in \left\{ 1,\dots, N_j\right\} },x _0^{(j)} , \boldsymbol{\theta}^{(j)}\right) \end{array}\right) \in \mathbb{R} ^{N_j}, \enspace j\in \left\{ 1, \dots, p\right\},
\end{equation}
where the polynomials $V^{(j)}_R$ are defined as in \eqref{V_R_polynomial}.
The assumption that the input signal $\left\{ {\bf z}  (t)\right\}_{t \in \Bbb Z }$ is a family of $n$-dimensional  independent and identically distributed random variables implies that the same property holds for the family $\left\{\boldsymbol{ \varepsilon } ^{( {\bf X} _0 ,\boldsymbol{\Theta} )}  (t)\right\}_{t \in \Bbb Z }$ of $N ^\ast  $-dimensional  random variables   in \eqref{linearization of the parallel reservoir}, namely, that $\left\{\boldsymbol{ \varepsilon } ^{( {\bf X} _0 ,\boldsymbol{\Theta} )}  (t)\right\}_{t \in \Bbb Z } \sim {\rm IID} (  \boldsymbol{\mu} ^{( {\bf X} _0 ,\boldsymbol{\Theta} )}_{\varepsilon }, \Sigma _{\varepsilon } ^{( {\bf X} _0 ,\boldsymbol{\Theta} )}) $. The mean  $\boldsymbol{\mu} ^{( {\bf X} _0 ,\boldsymbol{\Theta} )}_{\varepsilon }$ can be written as
\begin{equation}
\label{parallel reservoir mu_eps}
 \boldsymbol{\mu} _{{\varepsilon} }^{({\bf X} _0, \boldsymbol{\Theta} )}=\left( \begin{array}{c} \boldsymbol{\mu} _{{{\varepsilon} } }^{({\bf x} _0^{(1)}, \boldsymbol{\theta}^{(1)} )}\\\boldsymbol{\mu} _{{ {\varepsilon} } }^{({\bf x} _0^{(2)}, \boldsymbol{\theta} ^{(2)})}\\\vdots\\\boldsymbol{\mu} _{{ {\varepsilon} } }^{({\bf x} _0^{(p)}, \boldsymbol{\theta} ^{(p)})}\end{array} \right),
\end{equation}
where  $\boldsymbol{\mu} _{{{\varepsilon} } }^{({\bf x} _0^{(j)}, \boldsymbol{\theta}^{(j)} )} := {\rm } E \left[ \boldsymbol{\varepsilon} (t) ^{({\bf x} _0^{(j)}, \boldsymbol{\theta}^{(j)} )}\right]  \in \mathbb{R} ^{N_j}$, $j\in \left\{ 1, \dots, p\right\} $, whose components are determined by an expression  of the form~\eqref{mu_eps_res_main_r}, that is,
\begin{align}
\label{mu_eps_res_main_r parallel}
 \left(\boldsymbol{\mu} _{{\varepsilon} }^{({\bf x} _0^{(j)}, \boldsymbol{\theta} ^{(j)})}\right)_r= &(1 - { e} ^{- \xi^{(j)}})\sum^{R}_{i = 1}   (\partial _{{I} } ^{(i)} f^{(j)}   )({x} _0^{(j)} , 0, \boldsymbol{\theta} ^{(j)} ) \sum^{r}_{j' = 1} e^{-(r-j') \xi^{(j)}  }  \nonumber\\
 & \times  \sum^{}_{k _1 +\dots + k _n  = i} \dfrac{1}{k _1! \dots k _n!} \prod^{n}_{s=1} (C^{(j)}_{j's})^{k_s} \cdot \mu _{k _1 , \dots, k _n }({\bf z} ), \enspace r\in \left\{ 1, \dots, N_j\right\}. 
\end{align}
Additionally, the covariance matrix $\Sigma _{{ \varepsilon }}^{( {\bf X} _0 ,\boldsymbol{\Theta} )}:={\rm E}\left[ (\boldsymbol{ \varepsilon } ^{( {\bf X} _0 ,\boldsymbol{\Theta} )}(t)  - \boldsymbol{\mu}_{{ \varepsilon }}^{( {\bf X} _0 ,\boldsymbol{\Theta} )}) (\boldsymbol{ \varepsilon } ^{( {\bf X} _0 ,\boldsymbol{\Theta} )}(t)  - \boldsymbol{ \mu}_{{ \varepsilon }}^{( {\bf X} _0 ,\boldsymbol{\Theta} )}) ^\top \right]$  can be written as
\begin{equation}
\label{Sigma_eps_main parallel block}
\Sigma _{{ \varepsilon }}^{( {\bf X} _0 ,\boldsymbol{\Theta} )}= \left( \begin{array}{ccc} \Sigma _{{ \varepsilon }}^{( {\bf x}^{(1)} _0 ,\boldsymbol{\theta} ^{(1)} ),( {\bf x}^{(1)} _0 ,\boldsymbol{\theta} ^{(1)} ) }&\dots&\Sigma _{{ \varepsilon }}^{( {\bf x}^{(1)} _0 ,\boldsymbol{\theta} ^{(1)} ),( {\bf x}^{(p)} _0 ,\boldsymbol{\theta} ^{(p)} ) }\\
\vdots&\ddots&\vdots\\
\Sigma _{{ \varepsilon }}^{( {\bf x}^{(p)} _0 ,\boldsymbol{\theta} ^{(p)} ),( {\bf x}^{(1)} _0 ,\boldsymbol{\theta} ^{(1)} ) }&\dots&\Sigma _{{ \varepsilon }}^{( {\bf x}^{(p)} _0 ,\boldsymbol{\theta} ^{(p)} ),( {\bf x}^{(p)} _0 ,\boldsymbol{\theta} ^{(p)} ) }
\end{array} \right),
\end{equation}
where each block $\Sigma _{{ \varepsilon }}^{( {\bf x}^{(i)} _0 ,\boldsymbol{\theta} ^{(i)} ),( {\bf x}^{(j)} _0 ,\boldsymbol{\theta} ^{(j)} ) } \in \mathbb{M} _{N_i, N_j}$, $i, j \in \left\{ 1, \dots, p \right\} $ represents the covariance between the innovation components  that drive the $i$th and the $j$th time-delay reservoirs, respectively, and which has entries determined by:
\begin{align}
\label{Sigma_eps_main parallel}
(\Sigma _{{ \varepsilon }}^{( {\bf x}^{(i)} _0 ,\boldsymbol{\theta} ^{(i)} ),( {\bf x}^{(j)} _0 ,\boldsymbol{\theta} ^{(j)} ) }) _{rs} = &(1 - { e} ^{- \xi ^{(i)} }) (1 - { e} ^{- \xi ^{(j)}})  {\rm E}\Big[ V^{(i)}_R \left(  {\bf z} (t), \left\{ C^{(i)}_{j',\cdot}\right\}_{j' \in \left\{ 1, \dots, r\right\} },x _0 ^{(i)}, \boldsymbol{\theta}^{(i)} \right) \nonumber \\
& \times  V^{(i)}_R \left(  {\bf z} (t), \left\{ C^{(j)}_{j',\cdot}\right\}_{j' \in \left\{ 1, \dots, s\right\} },x _0^{(j)} , \boldsymbol{\theta} ^{(j)}\right)  \Big] \nonumber \\
&- (\boldsymbol{\mu} _{{\varepsilon} }^{({\bf x} _0^{(i)}, \boldsymbol{\theta} ^{(i)})})_r(\boldsymbol{\mu} _{{\varepsilon} }^{({\bf x} _0^{(j)}, \boldsymbol{\theta} ^{(j)})})_s, \enspace r \in \left\{ 1, \dots, N_i \right\}, \enspace s \in \left\{ 1, \dots, N_j \right\},
\end{align}
where the first summand is computed using the same approach that we described in expression~\eqref{Sigma_eps_main}.

The connectivity matrix can be easily written in terms of the connectivity matrices of each of the reservoirs that make up the parallel pool as
\begin{equation}
\label{A parallel reservoir}
A({\bf X} _0, \boldsymbol{\Theta}  ) := D_{\mathbf{X} } F({\bf X} _0, {\bf 0} _{N ^\ast  }\boldsymbol{\Theta} )=\left(\begin{array}{cccc} A^{(1)}({\bf x} _0^{(1)}, \boldsymbol{\theta}^{(1)} )& \mathbb{O}_{N_1,N _2}&\cdots& \mathbb{O}_{N_1, N _p} \\
 \mathbb{O}_{N_2, N _1}&A^{(2)}({\bf x} _0^{(2)}, \boldsymbol{\theta}^{(2)} )&\cdots&\mathbb{O}_{N_2, N _p}\\
 \vdots&\vdots&\ddots&\vdots\\
 \mathbb{O}_{N_p, N _1}&\mathbb{O}_{N_p, N _2}&\cdots&A^{(p)}({\bf x} _0^{(p)}, \boldsymbol{\theta}^{(p)} )\end{array}\right),
\end{equation}
 where for each $j\in \left\{ 1, \dots, p \right\} $ the matrix $A^{(j)}({\bf x} _0^{(j)},\boldsymbol{\theta} ^{(j)}):=D_{\mathbf{x} } F^{(j)}({\bf x} _0^{(j)}, {\bf 0} _{N_j}, \boldsymbol{\theta}^{(j)} )$  is the connectivity matrix of the $j$th TDR, determined as the first  derivative of the  corresponding $j$th reservoir map $F^{(j)}$ with respect to its first argument, evaluated at the point $ ({\bf x} _0^{(j)}, {\bf 0} _{N_j}, \boldsymbol{\theta}^{(j)} )$. Each of those individual connectivity matrices $A^{(j)}({\bf x} _0^{(j)},\boldsymbol{\theta} ^{(j)}) $  has the  explicit form provided in~\eqref{DxF}. Moreover, if for each of the individual equilibria $x _0^{(j)} $ used in the construction we require the stability condition $|\partial _{{x}}f^{(j)}({ x} ^{(j)}_0, { 0} , \boldsymbol{\theta}  )| < 1$, then by the proof of Theorem D.10 in~\cite{GHLO2014_capacity}  we have that the spectral radii $\rho(A^{(j)}(\mathbf{x}^{(j)} _0,  \boldsymbol{\theta}^{(j)} )) <1$ and, consequently 
\begin{equation}
\label{stability parallel}
\rho(A({\bf X} _0, \boldsymbol{\Theta}  )) <1.
\end{equation}
 
In these conditions, \eqref{linearization of the parallel reservoir} determines a VAR(1) model driven by the noise $\left\{\boldsymbol{ \varepsilon } ^{( {\bf X} _0 ,\boldsymbol{\Theta} )}  (t)\right\}_{t \in \Bbb Z } \sim {\rm IID} (  \boldsymbol{\mu} ^{( {\bf X} _0 ,\boldsymbol{\Theta} )}_{\varepsilon }, \Sigma _{{ \varepsilon }}^{( {\bf X} _0 ,\boldsymbol{\Theta} )} ) $ and that has a unique causal and  second order stationary solution $\left\{ {\bf X} (t)\right\} _{t\in \Bbb Z }$ with time-independent mean
\begin{equation}
\label{parallel reservoir mu_x}
 \boldsymbol{\mu} _{{X} }^{({\bf X} _0, \boldsymbol{\Theta} )}=\left( \begin{array}{c} \boldsymbol{\mu} _{{{x} } }^{({\bf x} _0^{(1)}, \boldsymbol{\theta}^{(1)} )}\\\boldsymbol{\mu} _{{ {x} } }^{({\bf x} _0^{(2)}, \boldsymbol{\theta} ^{(2)})}\\\vdots\\\boldsymbol{\mu} _{{ {x} } }^{({\bf x} _0^{(p)}, \boldsymbol{\theta} ^{(p)})}\end{array} \right),
\end{equation}
with 
\begin{equation}
\label{parallel reservoir mu_x jth reservoir}
\boldsymbol{\mu} _{{{x} } }^{({\bf x} _0^{(j)}, \boldsymbol{\theta}^{(j)} )} =  (\mathbb{I}  _{N_j} -  A^{(j)}(\mathbf{x} _0^{(j)},  \boldsymbol{\theta}^{(j)} ) )^{-1} (F^{(j)}(\mathbf{x} _0^{(j)} , {\bf 0} _{N_j} , \boldsymbol{\theta}^{(j)} ) - A^{(j)}(\mathbf{x} _0^{(j)},  \boldsymbol{\theta}^{(j)} )  \mathbf{x} _0^{(j)}  + \boldsymbol{\mu} _{{{\varepsilon} } }^{({\bf x} _0^{(j)}, \boldsymbol{\theta}^{(j)} )}).
\end{equation}
This allows us to write the parallel reservoir model \eqref{linearization of the parallel reservoir} in mean-adjusted form as
\begin{equation}
\label{VAR model reservoir array}
\mathbf{X} (t) - \boldsymbol{\mu} _{{X} }^{({\bf X} _0, \boldsymbol{\Theta} )} = A({\bf X} _0, \boldsymbol{\Theta}  )( \mathbf{X} (t-1) - \boldsymbol{\mu} _{{X} }^{({\bf X} _0, \boldsymbol{\Theta} )}) + (\boldsymbol{\varepsilon } (t)^{({\bf X} _0, \boldsymbol{\Theta} )} - \boldsymbol{\mu}_{{\varepsilon }}^{({\bf X} _0, \boldsymbol{\Theta} )}).
\end{equation}
Finally, the autocovariance function $\Gamma (k):= {\rm E}[(\mathbf{X}(t)- \boldsymbol{\mu} _{{X} }^{({\bf X} _0, \boldsymbol{\Theta} )}) (\mathbf{X}(t-k)- \boldsymbol{\mu} _{{X} }^{({\bf X} _0, \boldsymbol{\Theta} )})^\top]$  of $\left\{ {\bf X} (t)\right\} _{t\in \Bbb Z }$ at lag $k \in \Bbb Z $ is determined  by the Yule-Walker equations~\cite{luetkepohl:book} whose solutions in vectorized form are:
\begin{align}
\label{Gamma0 parallel} 
{\rm vech}(\Gamma(0) )&= \left(\mathbb{I}_{N ^{\ast  '}}-L_{N^\ast  }(A(\mathbf{X} _0,  \boldsymbol{\Theta} )\otimes A(\mathbf{X} _0,  \boldsymbol{\Theta} ))D_{N^\ast  }\right)^{-1} {\rm vech}(\Sigma _{{ \varepsilon }}^{( {\bf X} _0 ,\boldsymbol{\Theta} )}),\\
\label{Gammak paralel} 
\Gamma(k)&=A(\mathbf{X} _0,  \boldsymbol{\Theta} ) \Gamma(k-1) \enspace {\rm with} \enspace \Gamma(-k)= \Gamma(k)^\top,
\end{align}
where  $N^{ \ast '} := \dfrac{1}{2}N ^\ast   (N^\ast  +1)$ and $L_{N^\ast  } \in \mathbb{M} _{N^{ \ast '}, N^{\ast 2}}$, $D_{N^\ast  } \in   \mathbb{M} _{N^{\ast 2}, N^{ \ast '}}$ are the elimination and the duplication matrices, respectively.

\subsection{Memory capacity estimations for multidimensional memory tasks}
\label{Multidimensional memory tasks}

In what follows we explain how to use the reservoir models~\eqref{VAR model reservoir} and~\eqref{VAR model reservoir array} presented in the previous two sections as well as their dynamical features in order to explicitly compute the reservoir capacities~\eqref{capacity formula} associated to different memory tasks. As we explained in~\eqref{memory task def}, a memory task is determined by a map
\begin{align}
\begin{array}{cccc}
H: &\mathbb{R} ^{(h+1)n}&\longrightarrow &\mathbb{R}^q\\
&{\rm vec}( {\bf z}  (t) , \dots,  {\bf z}  (t - h)  )&\mapsto& \mathbf{y} (t),
\end{array}
\end{align}
with $q, h \in \mathbb{N} $, that is made out of $q$ different real valued functions of the input signal, $h$ time steps into the past. The reservoir memory capacity $
C_H( \boldsymbol{\theta}, C, \lambda)$ associated to $H$ measures the ability of the reservoir with parameters $\boldsymbol{\theta} $ to recover that function after being trained using a teaching signal.

In the next paragraphs we place ourselves in the context of a parallel array of $p$ time delay reservoirs as in Figure \ref{parallel reservoir figure}, with collective nonlinear kernel parameters $\boldsymbol{\Theta} \in \mathbb{R} ^{K^\ast} $ and operating in the neighbourhood of a stable fixed point ${\bf X} _0  \in \mathbb{R} ^{N ^\ast }$, with $N ^\ast  $ and $K ^\ast  $ the total number of neurons and the total number of parameters of the array, respectively. In order to estimate in this context the formula~\eqref{capacity formula}, we first use the parallel TDR model \eqref{VAR model reservoir array} in order to determine the autocovariance $\Gamma (0)  \in \Bbb S _{N ^\ast  }$ out of  the solutions~\eqref{Gamma0 parallel} of the Yule-Walker equation. The input signal $\left\{ {\bf z}  (t)\right\}_{t \in \Bbb Z }$ is assumed to be a  family of $n$-dimensional independent and identically distributed random variables with mean zero and  covariance matrix $\Sigma _z\in \Bbb S ^+ _n$, that is $\left\{ {\bf z}  (t)\right\}_{t \in \Bbb Z }\sim {\rm IID} ({\bf 0} _n , \Sigma _z )$. In these conditions, once a specific memory task of interest has been determined, all that is needed in order to complete the evaluation of the capacity formula~\eqref{capacity formula}  is the expressions for ${\rm Cov} (\mathbf{X} (t) , \mathbf{y} (t)) $ and ${\rm trace} \left({\rm Cov}(\mathbf{y} (t), \mathbf{y} (t))\right)$ that we now explicitly derive for the particular cases of the linear and quadratic memory tasks, respectively. 

\medskip

\noindent {\bf{Linear memory task.}}
Consider the linear $q$-dimensional $h$-lag memory task function $H: \mathbb{R} ^{(h+1)n} \longrightarrow \mathbb{R} ^q $ determined by the assignment $H( {\bf z} ^h (t) ) := L^\top  {\bf z} ^h (t) =:{\bf y}  (t) $, where ${\bf z} ^h (t) = {\rm vec}({\bf z}(t) ,{\bf z} (t-1), \dots,{\bf z} (t-h))  \in\mathbb{R} ^{(h+1)n}$ and  $L \in \mathbb{M} _{(h+1)n ,  q}$.  We now compute in this case  ${\rm Cov} ( \mathbf{X} (t)  ,\mathbf{y}  (t) )$ and ${\rm trace} \left({\rm Cov}(\mathbf{y} (t), \mathbf{y} (t))\right)={\rm Cov} ( \mathbf{y} (t) ^\top ,\mathbf{y} (t) ^\top )$, that are required for the memory  capacity evaluation.

\noindent{\bf (i)} We start with ${\rm Cov} ( \mathbf{y} (t) ^\top ,\mathbf{y} (t) ^\top )$ and write
\begin{align}
\label{}
{\rm Cov} ( \mathbf{y} (t) ^\top ,\mathbf{y} (t) ^\top )&= {\rm E} \left[\mathbf{y} (t) ^\top\mathbf{y} (t) \right] - {\rm E} \left[\mathbf{y} (t) ^\top \right] {\rm E} \left[\mathbf{y} (t)  \right]  \nonumber \\
&={\rm E} \left[  {\bf z} ^h (t)^\top L L^\top  {\bf z} ^h (t)  \right] = {\rm trace} \left(L L^\top{\rm E}[{\bf z} ^h (t){\bf z} ^h (t)^\top]\right)  =  {\rm trace}(LL^\top \Sigma _{{ z} ^h }),
 \end{align}
where the covariance matrix $\Sigma _{z^h} \in \mathbb{S} _{(h+1)n \times  (h+1)n} $ has the  form
\begin{equation*}
 \Sigma _{{ z} ^h } =  \left( \begin{array}{ccc} \Sigma _{{ z} } &\cdots& \mathbb{O} _{n}\\\vdots&\ddots&\vdots\\\mathbb{O} _{n}&\cdots&\Sigma _{{ z} } \end{array}\right).
\end{equation*}
\noindent{\bf (ii)} We now compute ${\rm Cov} ( \mathbf{X} (t),\mathbf{y}  (t) )$. As we already pointed out, the stability condition on the fixed point ${\bf X} _0 $ implies that the unique stationary solution of the ${\rm VAR}(1)$ model~\eqref{VAR model reservoir array} admits a ${\rm MA}(\infty)$ representation of the form: 
\begin{equation}
\label{}
\mathbf{X} (t) - \boldsymbol{\mu} _{{X} }^{({\bf X} _0, \boldsymbol{\Theta} )}   = \sum^{\infty}_{j = 0} \Psi _j  \boldsymbol{\rho} (t-j),
\end{equation}
with $\Psi _j\in \mathbb{M} _{N ^\ast}$ and $ \boldsymbol{\rho} (t) := \boldsymbol{\varepsilon } (t)^{({\bf X} _0, \boldsymbol{\Theta} )} - \boldsymbol{\mu}_{{\varepsilon }}^{({\bf X} _0, \boldsymbol{\Theta})}$.  Then, for any $i \in\left\{ 1, \dots, N ^\ast  \right\} $ and $j \in \left\{ 1, \dots, q \right\} $ we write
\begin{align}
\label{}
{\rm Cov}&(X_i (t) , y_j (t))=\sum^{\infty}_{k= 0} {\rm Cov} (( \Psi _k \boldsymbol{\rho} (t-k))_i, (L ^\top\cdot {\bf z}^h (t) )_j)
=\sum^{\infty}_{k= 0} \sum^{N ^\ast  }_{u = 1}\sum^{(h+1)n }_{v = 1} ( \Psi _k )_{iu} L_{vj} {\rm E} [\rho _u (t-k) ({\bf z}^h (t) )_v ] \nonumber\\
&= \sum^{\infty}_{k= 0} \sum^{N ^\ast  }_{u = 1}\sum^{n  }_{v = 1} \sum^{h+1}_{s = 1} ( \Psi _k )_{iu} L_{v\cdot s,j} {\rm E} [({\varepsilon } _u(t - k)^{({\bf X} _0, \boldsymbol{\Theta} )} - (\boldsymbol{\mu}_{{\varepsilon }}^{({\bf X} _0, \boldsymbol{\Theta})})_u) z_v(t-s+1) ]\nonumber\\
&=  \sum^{N ^\ast  }_{u = 1}\sum^{n  }_{v = 1} \sum^{h}_{s = 0}( \Psi _s )_{iu} L_{v\cdot s, j}  {\rm E} [{\varepsilon } _u(t - s)^{({\bf X} _0, \boldsymbol{\Theta} )} z _v(t - s) ],
\end{align} 
where the  vector $\boldsymbol{\varepsilon } (t)^{({\bf X} _0, \boldsymbol{\Theta} )}$ is provided in \eqref{parallel reservoir epsilon}-\eqref{eps_t_V jth reservoir} and the expectations ${\rm E} [{\varepsilon } _u(t - s)^{({\bf X} _0, \boldsymbol{\Theta} )} z _v(t - s) ] $ are computed by multiplying the $V _R $  polynomial corresponding to ${\varepsilon } _u(t)^{({\bf X} _0, \boldsymbol{\Theta} )} $ by the monomial $z _v(t )$; the resulting polynomial is the evaluated on the higher order moments of $\{ {\bf z} (t)\} _{t \in \Bbb Z}$ using the same rule that we stated after~\eqref{Sigma_eps_main}.

\medskip

\noindent {\bf {Quadratic memory task.}} Consider now the quadratic $q$-dimensional $h$-lag memory task function $H: \mathbb{R} ^{(h+1)n} \longrightarrow \mathbb{R} ^q $ determined by the assignment  $H( {\bf z} ^h (t) ) := Q \cdot {\rm vech}({\bf z} ^h (t) \cdot {\bf z} ^h (t) ^\top) =:{\bf y}  (t) $, where ${\bf z} ^h (t) = {\rm vec}({\bf z}(t) ,{\bf z} (t-1), \dots,{\bf z} (t-h))  \in\mathbb{R} ^{(h+1)n}$, $Q \in \mathbb{M} _{q ,  q ^\ast  }$ and $ q^\ast  := \dfrac{1}{2} (h+1)n((h+1)n+1)$. We  now provide explicit expressions for ${\rm Cov} ( \mathbf{X} (t)  ,\mathbf{y}  (t) )$ and ${\rm Cov} ( \mathbf{y} (t) ^\top ,\mathbf{y} (t) ^\top )$ in this case, that are required in order to evaluate  the corresponding memory capacity.

\noindent {\bf (i) } We start with ${\rm Cov}(\mathbf{y} (t) ^\top, \mathbf{y} (t) ^\top)$. Let $M ^h := {\bf z} ^h (t) {\bf z} ^h (t)^\top$ and write 
\begin{align}
\label{cov yy quadratic}
{\rm Cov}(\mathbf{y} (t) ^\top, \mathbf{y} (t) ^\top)&={\rm E} \left[\mathbf{y} (t) ^\top\mathbf{y} (t) \right] - {\rm E} \left[\mathbf{y} (t) ^\top \right] {\rm E} \left[\mathbf{y} (t)  \right] \nonumber\\
&= {\rm E}[({\rm vech}( {\bf z} ^h (t) {\bf z} ^h (t)^\top) )^\top Q^\top Q ({\rm vech}( {\bf z} ^h (t) {\bf z} ^h (t)^\top) )]- {\rm E} \left[\mathbf{y} (t) ^\top \right] {\rm E} \left[\mathbf{y} (t)  \right] \nonumber\\
&={\rm trace }(Q ^\top Q \cdot {\rm E} [ {\rm vech}(M ^h) ({\rm vech}(M ^h))^\top ])- {\rm E} \left[\mathbf{y} (t) ^\top \right] {\rm E} \left[\mathbf{y} (t)  \right].
\end{align}
Notice now that for any $i,j \in \left\{ 1, \dots, (h+1)n\right\} $ there exist $l _i , l _j \in \left\{ 0, \dots, h \right\} $ and $m _i , m _j\in \left\{ 0, \dots, n \right\} $ such that $i = l _i p + m_i $, $j = l _j p + m _j $ and hence   
\begin{equation}
\label{}
M ^h _{ij} = ({\bf z} ^h (t) {\bf z} ^h (t)^\top)_{ij}  = z _{m _i } (t - l_i ) z _ {m _j } (t - l _j ).
\end{equation}
Consequently, for any $i,j \in \left\{ 1, \dots, (h+1)n\right\} $
\begin{align}
\label{EMhMh quadratic}
{\rm E} [{\rm vech}(M^h ) {\rm vech}(M^h )^\top  _{ij}] &= {\rm E } [ {\rm vech}(M ^h ) _i {\rm vech}(M^h) _j   ]= {\rm E} [M ^h _{ \sigma^{-1} (i)} M ^h _{ \sigma^{-1} (j)}] \nonumber\\
&= {\rm E} [z _{ m _{r(i)} } (t - l _{r(i)} ) z _{m _{s (i)}} (t - l _{s(i)} )z_{m_{u(j)}} (t- l _{u(j)} ) z _{m _{v(j)} } (t - l _{v(j)} )],
\end{align}
where the operator $\sigma^{-1}$ assigns to the index of the position of an  element in ${\rm vech}(M^h)$ the two indices corresponding to its position in the matrix $M^h \in \Bbb S _{q^\ast  }$. In this expression $(r(i),s(i)) := \sigma ^{-1} (i)$, $(u(j),v(j)) := \sigma^{-1} (j) $ and $ r(i) = l _{r(i)} p+ m _{r(i)} $, ${s(i)}=l _{s(i)} p + m _{s(i)} $, $u(j) = l _{u(j)} p + m _{u(j)} $, $v(j) = l _{v(j)} p + m _{v(j)} $ with $l _{r(i)} , l _{s(i)} , l _{u(j)}, l _{v(j)} \in \left\{ 0, \dots, h\right\} $ and $m _{r(i)} , m _{s(i)} , m _{u(j)} , m _{v(j)} \in \left\{ 1, \dots, p \right\} $. Additionally, using this notation the following relation holds true
\begin{equation}
\label{Ey quadratic}
{\rm E} \left[ {y} _k(t) \right] = \sum^{ q ^\ast  }_{i = 1}  Q_{kj} {\rm E}[( {\rm vech}(M ^h )) _i  ] = \sum^{ q^\ast  }_{j = 1} Q_{kj} {\rm E} [ z_{m_{r(i)}} (t-l_{r(i)})z_{m_{s(i)}} (t-l_{s(i)}) ].
\end{equation}
We hence now can  derive the expression for \eqref{cov yy quadratic} as
\begin{align}
\label{cov yy quadratic final}
{\rm Cov}(\mathbf{y} (t) ^\top, \mathbf{y} (t) ^\top)&=\sum^{ q^\ast  }_{i = 1}\sum^{q}_{k = 1}   Q_{ki}\Big(\sum^{q ^\ast  }_{j= 1}  Q_{kj} \cdot {\rm E} [z _{ m _{r(i) }} (t - l _{r(i) } ) z _{m _{s(i) } } (t - l_{s(i) } )z_{m_{u(j)}} (t- l _{u(j)} ) z _{m _{v(j)} } (t - l _{v(j)} )]\nonumber\\
&-  Q_{ki} {\rm E} [ z_{m_{r(i)}} (t-l_{r(i)})z_{m_{s(i)}} (t-l_{s(i)}) ]^2\Big),
\end{align}
where we used the same notation as in \eqref{EMhMh quadratic} and \eqref{Ey quadratic}.

\noindent{\bf (ii) } ${\rm Cov} ( \mathbf{X} (t)  ,\mathbf{y}  (t) )$: for any $i \in \left\{ 1, \dots, N ^\ast  \right\} $, $j\in \left\{ 1, \dots, q \right\} $ we have
\begin{align}
\label{}
{\rm Cov}&(X_i (t) , y_j (t))=\sum^{\infty}_{k= 0} {\rm Cov} (( \Psi _k \boldsymbol{\rho} (t-k))_i, (Q\cdot {\rm vech}(M^h ) )_j)\nonumber\\
&=\sum^{\infty}_{k= 0} \sum^{N^\ast  }_{u = 1}\sum^{q ^\ast  }_{v = 1} ( \Psi _k )_{iu} Q_{jv} {\rm E} [\rho _u (t-k) ({\rm vech}(M ^h ) )_v ] \nonumber\\
&= \sum^{\infty}_{k= 0} \sum^{N ^\ast  }_{u = 1}\sum^{q ^\ast  }_{v = 1} ( \Psi _k )_{iu} Q_{jv} {\rm E} [({\varepsilon } _u(t - k)^{({\bf X} _0, \boldsymbol{\Theta} )}- (\boldsymbol{\mu} _ \varepsilon ^{({\bf X} _0, \boldsymbol{\Theta} )})_u) ({\rm vech}(M ^h ) )_v ]\nonumber\\
&=\sum^{\infty}_{k= 0} \sum^{N^\ast  }_{u = 1}\sum^{q ^\ast  }_{v = 1} ( \Psi _k )_{iu} Q_{jv} \{ {\rm E} [({\varepsilon } _u(t - k)^{({\bf X} _0, \boldsymbol{\Theta} )}- (\boldsymbol{\mu} _ \varepsilon ^{({\bf X} _0, \boldsymbol{\Theta} )})_u) z _{m_{r(v)}}(t - l_{m_{r(v)}})z _{m_{s(v)}}(t - l_{m_{s(v)}}) ]\}\nonumber\\
&= \sum^{h}_{k= 0} \sum^{N^\ast  }_{u = 1}\sum^{q ^\ast  }_{v = 1} ( \Psi _k )_{iu} Q_{jv} \{ {\rm E}[ {\varepsilon } _u(t - k)^{({\bf X} _0, \boldsymbol{\Theta} )} \cdot z _{m_{r(v)}}(t - l_{m_{r(v)}})z _{m_{s(v)}}(t - l_{m_{s(v)}})]\nonumber\\
&\quad -(\boldsymbol{\mu} _ \varepsilon ^{({\bf X} _0, \boldsymbol{\Theta} )})_u {\rm E}[z _{m_{r(v)}}(t - l_{m_{r(v)}})z _{m_{s(v)}}(t - l_{m_{s(v)}})]\},
\end{align} 
where the vector $\boldsymbol{\varepsilon } (t)^{({\bf X} _0, \boldsymbol{\Theta} )}$ is provided in \eqref{parallel reservoir epsilon}-\eqref{eps_t_V jth reservoir} and where  we used the same notation as in \eqref{EMhMh quadratic} and \eqref{Ey quadratic}.

 \subsection{The impact of the  teaching signal size in the reservoir performance}
 \label{Finite sample effect in the Tikhonov regularised reservoir regression} 
 
In the preceding  sections we evaluated the reservoir characteristic error or, equivalently, its capacity, in terms of various second order moments of the input signal and the reservoir output that in turn can be explicitly written in terms of the reservoir parameters. There are situations in which those moments are obtained directly out of finite sample realizations of the teaching signal and of the reservoir output using empirical estimators. That approach introduces an estimation error in the readout layer $({ W}_{{\rm out}}, \boldsymbol{a}_{\rm out}) $ that 
 that adds to the characteristic error~\eqref{error rc optimal}. The quantification of that error is the main goal of this section. Since this error depends on the value of the regularizing constant $\lambda$, the pair $(W_\lambda, \boldsymbol{a}_\lambda) $ will denote  in what follows  the optimal readout layer $({ W}_{{\rm out}}, \boldsymbol{a}_{\rm out}) $ given by~\eqref{linear_cov_system_gamma}-\eqref{linear_cov_system_c} for a fixed value of the parameter $\lambda$.
 
 \medskip
 
\noindent {\bf Properties of the ridge estimator.} Consider now $\left\{ \boldsymbol{x} (1), \boldsymbol{x} (2), \dots, \boldsymbol{x} (T)\right\} $ and $\left\{ \boldsymbol{y} (1), \boldsymbol{y} (2), \dots, \boldsymbol{y} (T)\right\} $  samples of size $T$ of the reservoir output  $ \left\{ \boldsymbol{x} (t) \right\} _{t \in \mathbb{N} }$ and the teaching signal $ \left\{ \boldsymbol{y} (t) \right\} _{t \in \mathbb{N} }$ processes. We concatenate horizontally these observations and we obtain the matrices $X:=\left( \boldsymbol{x} (1)| \boldsymbol{x} (2)| \dots| \boldsymbol{x} (T)\right) \in \mathbb{M} _{N,T} $ and $Y:=\left( \boldsymbol{y} (1)| \boldsymbol{y} (2)| \dots| \boldsymbol{y} (T)\right) \in \mathbb{M} _{q,T}$. 
We now quantify the cost in terms of memory  capacity or, equivalently,  increased error, of using in the RC not the optimal readout layer $({ W}_\lambda, \boldsymbol{a}_\lambda) $ given by~\eqref{linear_cov_system_gamma}-\eqref{linear_cov_system_c} but an  estimation $(\widehat{W} _\lambda, \widehat{\boldsymbol{a} }_\lambda) $ of it based on estimators of the different moments present in those expressions using $X $ and $Y  $. More specifically, for a fixed $\lambda $ and samples $X $ and $Y  $, we produce an estimation $(\widehat{W} _\lambda, \widehat{\boldsymbol{a} }_\lambda) $ of $({ W}_\lambda, \boldsymbol{a}_\lambda) $ by using in~\eqref{linear_cov_system_gamma}-\eqref{linear_cov_system_c} the empirical estimators
\begin{align}
\label{}
\widehat{\boldsymbol{\mu} }_x &:= \dfrac{1}{T} X \mathbf{i} _T ,\qquad
\widehat{\boldsymbol{\mu} }_y := \dfrac{1}{T} Y \mathbf{i} _T ,\qquad
\widehat{ \Gamma} (0):= \dfrac{1}{T} XX^\top - \widehat{\boldsymbol{\mu} }_x\widehat{\boldsymbol{\mu} }_x^\top = \dfrac{1}{T}XAX^\top,
\label{estim1}\\
\reallywidehat{{\rm Cov} \left(\mathbf{y} (t), \mathbf{y} (t)\right)}&= \dfrac{1}{T} YY^\top - \widehat{\boldsymbol{\mu} }_y\widehat{\boldsymbol{\mu} }_y^\top = \dfrac{1}{T}YAY^\top,\label{estim2}\\
\reallywidehat{{\rm Cov} \left(\mathbf{x} (t), \mathbf{y} (t)\right)}  &=   \dfrac{1}{T} XY^\top - \widehat{\boldsymbol{\mu} }_x\widehat{\boldsymbol{\mu} }_y^\top = \dfrac{1}{T}XAY^\top,\label{estim3}
\end{align}
where $A:= \mathbb{I} _T - \dfrac{1}{T} \mathbf{i} _T\mathbf{i} _T ^\top$. These expressions substituted in~\eqref{linear_cov_system_gamma}-\eqref{linear_cov_system_c} yield
\begin{align}
\label{W hat finite sample}
\widehat{W} _\lambda =&(\widehat{ \Gamma} (0)+ \lambda \mathbb{I} _N )^{-1} \reallywidehat{{\rm Cov} \left(\mathbf{x} (t), \mathbf{y} (t)\right)} = (XAX^\top + \lambda T \mathbb{I}_N )^{-1} XAY^\top,\\
\widehat{\boldsymbol{a}}_\lambda =&\widehat{\boldsymbol{\mu} }_y   - \widehat{W}_\lambda ^\top \widehat{\boldsymbol{\mu} }_x = \dfrac{1}{T}(Y -\widehat{W}_\lambda ^\top X ) \mathbf{i} _T, \label{a hat finite sample}
\end{align}
and determine a finite sample  ridge regression estimator. The error associated to its use can be read out of its statistical properties that have been studied in the literature under various hypotheses. In the sequel we follow~\cite{GO13}, where  $(\widehat{W}_\lambda, \widehat{\boldsymbol{a}}_\lambda )$ is considered as a ridge estimator of the regression model 
\begin{align}
\label{regression}
\boldsymbol{y} (t) = \boldsymbol{a} + W^\top \boldsymbol{x} (t) + \boldsymbol{\varepsilon } (t), 
\end{align}
with $ \boldsymbol{a} \in \mathbb{R} ^q $ a constant vector-intercept, $W \in \mathbb{M} _{N, q}$, and where the different random variables in~\eqref{regression} are assumed to satisfy the following hypotheses:
\begin{description}
\item  [{\bf (H1)}] The error terms $\left\{ \boldsymbol{ \varepsilon } (t)\right\}_{t\in \Bbb Z}$ constitute a family of $q$-dimensional independent normally distributed vectors with mean ${\bf 0} _q$ and covariance matrix $\Sigma _{\varepsilon }^q \in \Bbb S _q $, that is, 
$\left\{ \boldsymbol{ \varepsilon } (t)\right\}_{t\in \Bbb Z} \sim {\rm IN}({\bf 0} _q , \Sigma _{\varepsilon }^q)$.
\item  [{\bf (H2)}] $\left\{ \mathbf{x} (t) \right\}_{t\in \Bbb Z}$ is a stochastic process such that $\boldsymbol{x} (t) \in \mathbb{R} ^N $ is a random vector independent of $\boldsymbol{ \varepsilon } (s) \in \mathbb{R} ^q $ for any $t, s\in \mathbb{Z} $.
\item  [{\bf (H3)}] The processes $\left\{ \mathbf{x} (t) \right\}_{t\in \Bbb Z}$, $\left\{ \mathbf{y} (t) \right\}_{t\in \Bbb Z}$, and the joint process $\left\{ (\mathbf{x} (t), \mathbf{y} (t) )\right\}_{t\in \Bbb Z}$ are second-order ergodic. This implies that the estimators in~\eqref{estim1}-\eqref{estim3} converge to the corresponding moments when $T \rightarrow \infty$ and also that these are second-order stationary (their means and autocovariances are time-independent).
\end{description}
Under those hypotheses, it can be shown that the estimator $(\widehat{W}_\lambda, \widehat{\boldsymbol{a}}_\lambda )$ has the properties described in  the following result whose proof can be found in~\cite{GO13}. In the statement we use the notation $Z\sim {\rm MN} _{m,n}(M_Z, U_Z, V_Z)$ to indicate that $Z\in \Bbb M _{m,n}$ is a matrix random variable distributed according to the matrix normal distribution with mean matrix $M_Z\in \Bbb M _{m,n}$ and   scale matrices   $U_Z\in \Bbb S _{m}$, $V_Z\in \Bbb S _{n}$. Details on these distributions can be found in~\cite{bookMatrixDistributions2000}, and references therein.

\begin{proposition}
\label{Properties of the ridge estimated coefficient matrix and the intercept vector}
 Consider the regression problem in~\eqref{regression} subjected to the hypotheses {\bf (H1)}-{\bf (H3)}  and let $\widehat{W} _\lambda \in \mathbb{M} _{N,q}$ and $\widehat{\boldsymbol{a}} _\lambda  \in \mathbb{R} ^q $ be the ridge estimators given by~\eqref{W hat finite sample} and~\eqref{a hat finite sample}, respectively, based on the samples of length $T$ contained in the matrices $X\in \mathbb{M} _{N,T} $ and $Y\in \mathbb{M} _{q,T}$. Then,
\begin{itemize}
 \item[\normalfont{\bf (i)}]  The distribution of the ridge estimator $\widehat{W} _\lambda|X $ of $W _\lambda $ conditioned by $X$ is given by
\begin{align}
\label{W normal}
(\widehat{W}_\lambda - W_\lambda) |X&\sim {\rm MN}_{N,q}(  - \lambda T R {W}_\lambda, \Sigma _{W}^r, \Sigma _{W}^c),
\end{align}
or, equivalently, 
\begin{align}
\label{W normal vectorial}
{\rm vec}(\widehat{W}_\lambda - W_\lambda) |X&\sim {\rm N}(  - \lambda T{\rm vec}(R {W}_\lambda)  , \Sigma _{W}^c \otimes \Sigma _{W}^r),
\end{align}
where the symbol $\otimes$ stands for the Kronecker product and where the row and column  covariance matrices $\Sigma _{W}^r\in \Bbb S _N $ and  $\Sigma _{W}^c \in \Bbb S _T $, respectively, are given by
\begin{equation}
\label{cov W}
\Sigma _{W}^r =  (\mathbb{I} _N - \lambda TR)R \quad \mbox{and} \quad
\Sigma _{W}^c =\Sigma_{\varepsilon }^q,
\end{equation}
where $R:=(XAX^\top+ \lambda T\mathbb{I} _N )^{-1}$ and  $A:=\mathbb{I}_T -\dfrac{1}{T}\mathbf{i} _T  \mathbf{i} _T ^\top$.
 \item[\normalfont{\bf (ii)}] The distribution of the ridge estimator $\widehat{\boldsymbol{a}} _\lambda |X$  of $\boldsymbol{a} _\lambda $  conditioned by $X$ is given by
\begin{align}
\label{a normal}
(\widehat{\boldsymbol{a}}_\lambda -\boldsymbol{a}_\lambda)|X &\sim {\rm N}(  \lambda {W}_\lambda^\top R X \mathbf{i} _T , \Sigma _{a}),
\end{align} 
where the  covariance matrix $\Sigma _a\in \Bbb S _q $ is given by
 \begin{align}
\label{cov a}
\Sigma _a:= {\dfrac{1}{T}}\left(  1+     {\dfrac{1}{T}}{\rm trace}((\mathbb{I} _N - \lambda TR)R X\mathbf{i} _T \mathbf{i} _T^\top X^\top) \right)  \Sigma _{\varepsilon}^q,
\end{align}
and where $R$ is defined as in {\bf (i)}.
\item[\normalfont{\bf (iii)}] The covariance $\Sigma _{Wa} \in \Bbb M _{Nq,q} $ between the conditioned ridge estimators ${\rm vec}(\widehat{W} _\lambda|X) \in \mathbb{R} ^{Nq}$ and $ \widehat{\boldsymbol{a}} _\lambda|X \in\mathbb{R} ^q $ is given by
 \begin{align}
\label{Sigma Wa}
\Sigma _{Wa}:= \dfrac{1}{T} ({\rm vec}(( \mathbb{I} _N - \lambda T R) W _\lambda ) \mathbf{i} _T ^\top X^\top ((\mathbb{I}_N - \lambda T R) W _\lambda) - \Sigma _{\varepsilon}^q \otimes( (\mathbb{I} _N - \lambda T R)R X\mathbf{i} _T )).
\end{align}
 \end{itemize}
 \end{proposition}

\medskip

\noindent {\bf The regression error with estimated regression parameters.} The reservoir error and the corresponding memory capacity in relations~\eqref{error rc optimal} and~\eqref{capacity formula}, respectively, were computed assuming that the optimal ridge parameters $(W _\lambda, {\bf a} _\lambda) $ are  known. This error, that is exclusively associated to the ability of the reservoir to perform or not the memory task in question, will be referred to as the {\bf characteristic error} and will be denoted as
\begin{multline}
\label{char error}
{\rm MSE}_{{\rm char, \lambda}}:= {\rm trace} \left({\rm E} \left[  ({ W_\lambda} ^{ \top} \cdot {\bf x}(t)  +\boldsymbol{a}_\lambda -\mathbf{y} (t)  ) ({ W_\lambda} ^{ \top} \cdot {\bf x}(t)  +\boldsymbol{a}_\lambda -\mathbf{y} (t)  )^\top\right]\right)\\
	= {\rm trace}({\rm Cov} \left(\mathbf{y} (t), \mathbf{y} (t)\right) -{\rm Cov} (\mathbf{x} (t), \mathbf{y} (t) )^\top(\Gamma (0) + \lambda \mathbb{I}  _N ) ^{-1} (\Gamma (0) + 2\lambda \mathbb{I} _N) (\Gamma (0) + \lambda \mathbb{I}  _N ) ^{-1}{\rm Cov} (\mathbf{x} (t), \mathbf{y} (t) )).
\end{multline}
When the reservoir is put to work using instead a readout layer $(\widehat{W}_\lambda, \widehat{\boldsymbol{a}}_\lambda )$ that has been estimated using finite sample realizations of the processes $\left\{ \mathbf{x} (t) \right\}_{t\in \Bbb Z}$ and $\left\{ \mathbf{y} (t) \right\}_{t\in \Bbb Z}$, the estimation error piles up with the characteristic one. The resulting error  will be called {\bf total error} and denoted ${\rm MSE}_{{\rm total}, \lambda}$. It can computed using the distribution properties of the ridge estimator spelled out in Proposition~\ref{Properties of the ridge estimated coefficient matrix and the intercept vector} under the assumption that the samples $(X,Y)$  that have been used to obtain the estimate $(\widehat{W}_\lambda, \widehat{\boldsymbol{a}}_\lambda )$ of the output layer and those to evaluate the RC performance are independent. Indeed, under those hypotheses, the total reservoir error ${\rm MSE}_{{\rm total}, \lambda}|X$ conditional on $X$ is
\begin{equation*}
\!\!\!\!\!\!\!\!\!
{\rm MSE}_{{\rm total, \lambda}}|X:= {\rm trace} \left({\rm E} \left[  ({ (W_\lambda+M _\lambda)} ^{ \top} \cdot {\bf x}(t)  +\boldsymbol{a}_\lambda+ {\bf v} _\lambda -\mathbf{y} (t)  ) ({ (W_\lambda+M _\lambda)} ^{ \top} \cdot {\bf x}(t)  +\boldsymbol{a}_\lambda+ {\bf v} _\lambda -\mathbf{y} (t)  )^\top|X\right]\right),
\end{equation*}
where the conditional  random variables $(M _\lambda, {\bf v} _\lambda)|X\in \mathbb{M} _{N,q} \times \mathbb{R} ^q$ are independent from the processes $\left\{ \mathbf{x} (t) \right\}_{t\in \Bbb Z}$ and $\left\{ \mathbf{y} (t) \right\}_{t\in \Bbb Z}$ and have the same distribution properties than those spelled out in Proposition~\ref{Properties of the ridge estimated coefficient matrix and the intercept vector} for $(\widehat{W}_\lambda-W _\lambda, \widehat{\boldsymbol{a}}_\lambda-\boldsymbol{a}_\lambda )|X$. The main result in~\cite{GO13} shows that, under those hypotheses, the total error conditional on $X$ is given by
\begin{eqnarray}
\label{MSE total theorem statement}
\!\!\!\!\!\!\!\!\!\!\!\!{\rm MSE}_{{\rm total, \lambda}}|X&= &{\rm MSE}_{{\rm char}, \lambda}  +
\dfrac{1}{T} {\rm trace }(\Sigma _{\varepsilon}^q)  + {\rm trace }(\Sigma _{\varepsilon}^q){\rm trace}\left( (\mathbb{I} _N - \lambda T\mathcal{R})\mathcal{R} Q \right) 
\nonumber\\
&+&  {\lambda ^2 }{T ^2 }{\rm trace}\left( W _\lambda ^\top \mathcal{R}Q \mathcal{R}W _\lambda  \right) +2\lambda^2 T  {\rm trace}(W _\lambda ^\top \mathcal{R} W _\lambda ),
\end{eqnarray}
where
\begin{align}
\label{}
Q:=& \Gamma (0) + \boldsymbol{\mu} _x \boldsymbol{\mu} _x ^\top + \dfrac{1}{T^2} X\mathbf{i} _T \mathbf{i} _T^\top  X^\top- \dfrac{2}{T} \boldsymbol{\mu} _x  \mathbf{i} _T ^\top X^\top 
\end{align}
and  ${\rm MSE}_{{\rm char}, \lambda} $ is given by~\eqref{char error}, ${ W}_{\lambda} = (\Gamma (0) + \lambda \mathbb{I}  _N ) ^{-1}{\rm Cov} (\mathbf{x} (t), \mathbf{y} (t) ) $,  and where
\begin{equation}
\label{important quantities}
R=(T\Gamma (0) + \lambda T\mathbb{I} _N )^{-1},\quad
\mathcal{R}=(XAX^\top+ \lambda  T\mathbb{I} _N )^{-1},\quad \mbox{and} \quad
A= \mathbb{I} _T - \dfrac{1}{T} \mathbf{i} _T \mathbf{i} _T ^\top.
\end{equation}
In empirical applications, the moments $\boldsymbol{\mu} _x $ and  $\Gamma (0) $ are in practice estimated with the same sample $X,Y $ used to evaluate the reservoir performance. Therefore, if we replace  $\boldsymbol{\mu} _x $,  $\Gamma (0) $, ${\rm Cov} \left(\mathbf{y} (t), \mathbf{y} (t)\right) $, and ${\rm Cov} \left(\mathbf{x} (t), \mathbf{y} (t)\right)$ in~\eqref{MSE total theorem statement} by ${\dfrac{1}{T}}X\mathbf{i} _T $,  $\dfrac{1}{T}XAX^\top$, $\dfrac{1}{T}YAY^\top$ and $\dfrac{1}{T}XAY^\top$, respectively,  we obtain the approximated but simpler expression ${\rm MSE}_{{\rm total}, \lambda}^{{\rm approx}} (\lambda)|X$ for the total error:
\begin{multline}
\label{almost final error}
{\rm MSE}_{{\rm total}, \lambda} ^{{\rm approx}}|X= {\rm MSE}_{{\rm char}, \lambda} ^{{\rm approx}}\\
+ \dfrac{1}{T}{\rm trace}(\Sigma _{\varepsilon} ^q)\bigg[ 1 + {\rm trace}((\mathbb{I} _N - \lambda T \mathcal{R}) \mathcal{R} XAX^\top)\bigg] +  \lambda ^2T{\rm trace}\left(W _\lambda ^\top  \mathcal{R} (3 \mathbb{I} _N  -\lambda T   \mathcal{R} )W_\lambda \right),
\end{multline}
with 
\begin{equation}
\label{}
{\rm MSE}_{{\rm char}, \lambda} ^{{\rm approx}}= \dfrac{1}{T }{\rm trace} \left( YAY^\top - YAX^\top  (\mathbb{I} _N + \lambda T \mathcal{R})\mathcal{R} XAY^\top \right).
\end{equation}
This approximation can be taken one step further by replacing in~\eqref{almost final error} ${ W}_{\lambda} = (\Gamma (0) + \lambda \mathbb{I}  _N ) ^{-1}{\rm Cov} (\mathbf{x} (t), \mathbf{y} (t) ) $ by its estimator  $\widehat{W} _\lambda  = (XAX^\top + \lambda T \mathbb{I}_N )^{-1} XAY^\top $ and $\Sigma _{\varepsilon} ^q $ by its natural estimator in terms of $X $ and $Y $ for which
\begin{equation*}
{\rm trace}(\widehat{\Sigma _{\varepsilon} ^q})= {\rm MSE}_{{\rm char}, \lambda} ^{{\rm approx}}.
\end{equation*}
With these two additional substitutions, the equality~\eqref{almost final error} yields the following approximated expression for the total error:
\begin{multline*}
{\rm MSE}_{{\rm total}, \lambda} ^{{\rm approx}}|X= {\rm MSE}_{{\rm char}, \lambda} ^{{\rm approx}}+\\
+ \dfrac{1}{T} {\rm MSE}_{{\rm char}, \lambda} ^{{\rm approx}}\bigg[ 1 + {\rm trace}((\mathbb{I} _N - \lambda T \mathcal{R}) \mathcal{R} XAX^\top)\bigg] +  \lambda ^2T{\rm trace}\left(Y A X^\top   \mathcal{R}^2 (3 \mathbb{I} _N  -\lambda T   \mathcal{R} )\mathcal{R} XAY^\top \right).
\end{multline*}

 \section{Empirical study}
\label{Empirical study}
 
 The goal of this section is twofold. First, we will assess the ability of the reservoir model in the multidimensional setup introduced in Section~\ref{The approximated model of TDR with multidimensional input signals} to produce in that context good estimates of the memory capacity of the original reservoir, both in the continuous and the discrete-time setups. Our study shows that there is a good match between the performances of the model and of the actual reservoir and hence shows that the explicit expression of the reservoir capacity coming from the model can be used to find, for a given multidimensional  task with multidimensional input signal, parameters and input masks for the original system that optimize its performance.
 
 Second, we will use the parallel reservoir model in Section~\ref{The approximating model of parallel TDRs with multidimensional input signals} and the memory capacity formulas that can be explicitly written with it, in order to verify various robustness properties of this reservoir architecture that were already documented in \cite{GHLO2012}. These properties have to do mainly with task misspecification and sensitivity to the parameter choice.
 
 \medskip
 
\noindent {\bf {Evaluation of the TDR performance in the processing of multidimensional signals.}} 

\medskip

\begin{figure}[h!]
\hspace{-2.4cm}
\includegraphics[scale=.5, angle=0]{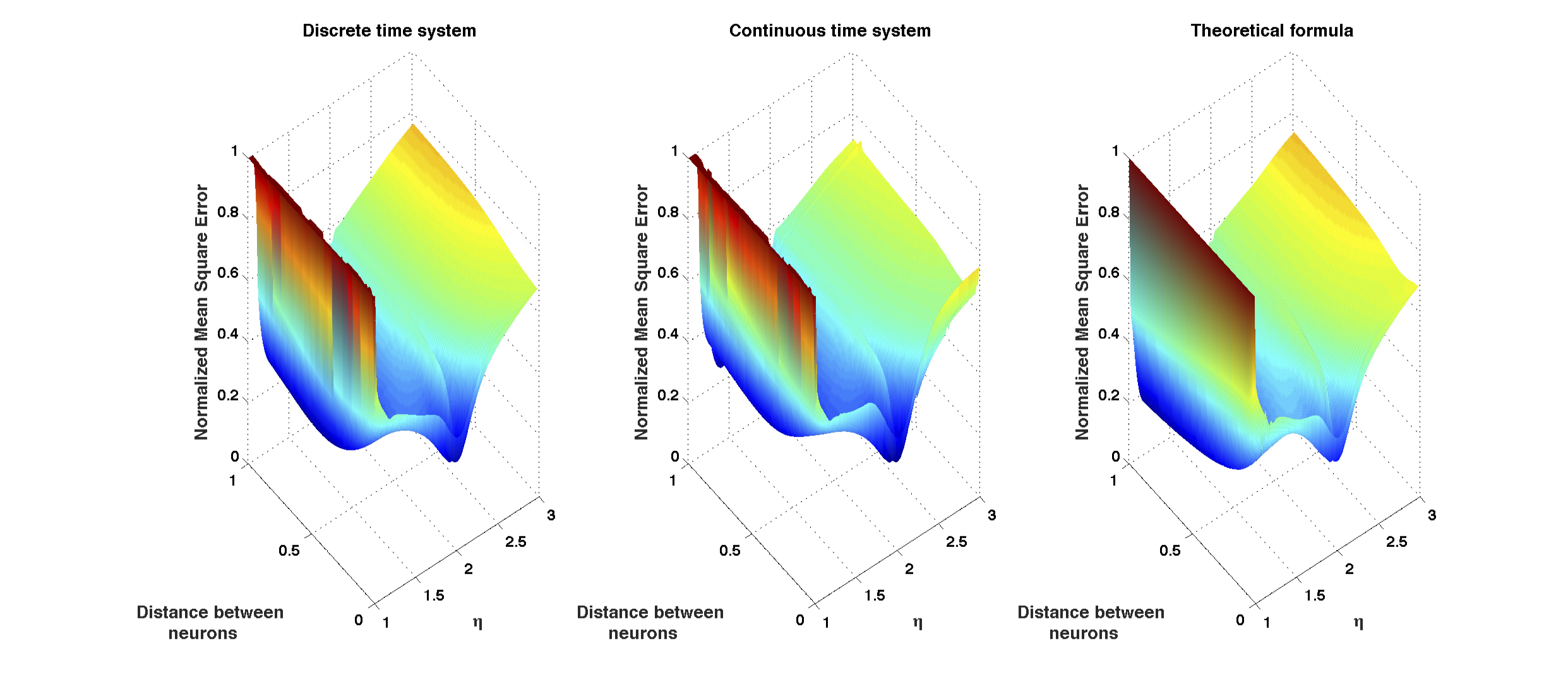}
\caption{Normalized mean square error surfaces exhibited by an individually operating TDR constructed using a nonlinear Mackey-Glass kernel \eqref{Mackey-Glass nonlinear kernel} performing a quadratic 3-lags memory task on a 3-dimensional independent  mean zero input signal with  covariance matrix $\Sigma _z$ given by  ${\rm vech}(\Sigma _z ) = (0.0016, 0.0012, 0.0008, 0.0017, 0.0002, 0.0018)$. In these figures the input gain $ \gamma = 0.6163$ is  kept constant. The values of the input mask $C\in \mathbb{M} _{20, 3}$ are chosen randomly using a uniform distribution in the interval $\left[ -1; 1 \right] $. The left and middle panels show the error surfaces exhibited by the discrete and continuous time reservoirs computed via Monte Carlo simulations. The right panel shows the error produced by the reservoir model and computed evaluating  the explicit capacity formula that can be written down in that case.}
\label{multidimensional MG}
\end{figure}

\begin{figure}[h!]
\hspace{-2.4cm}
\includegraphics[scale=.5, angle=0]{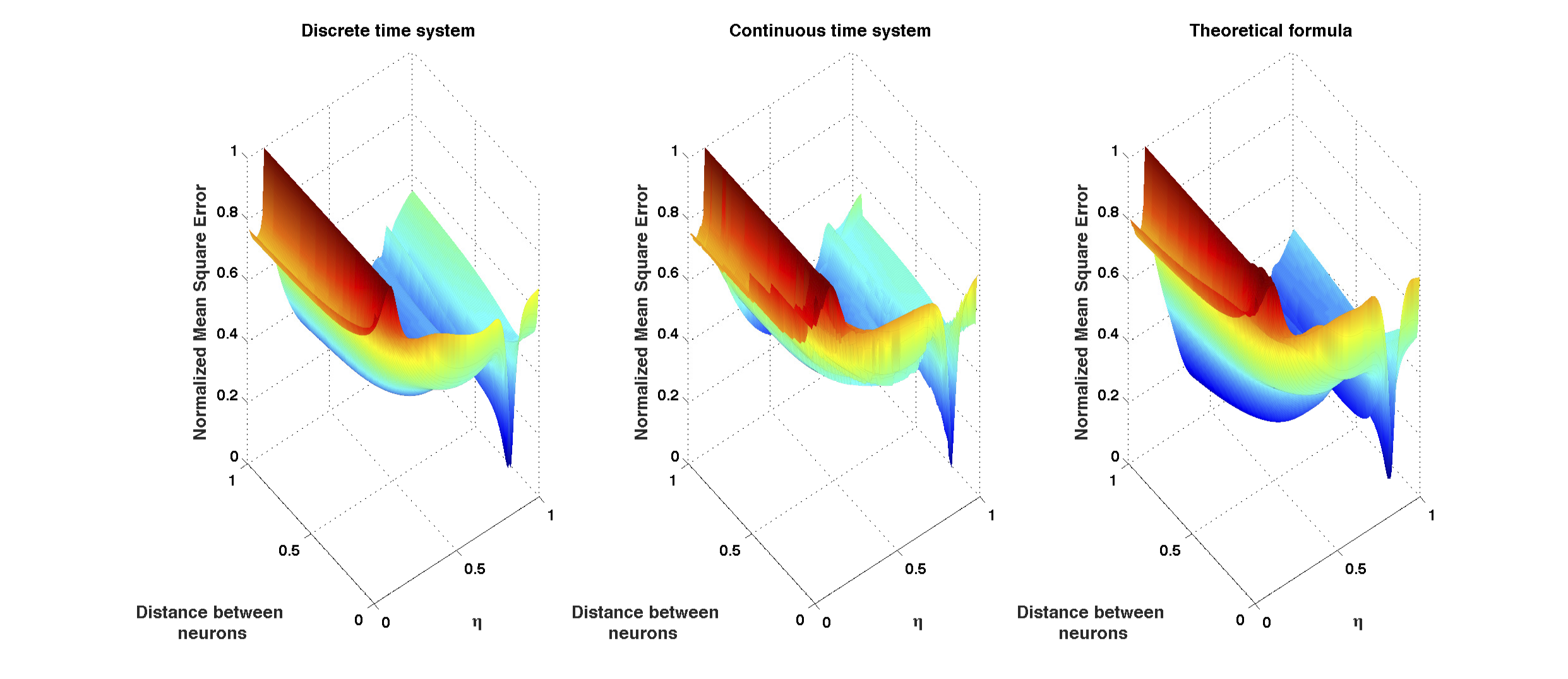}
\caption{Normalized mean square error surfaces exhibited by an individually operating TDR constructed using a nonlinear Ikeda kernel \eqref{Ikeda nonlinear kernel} performing a quadratic 3-lags memory task on a 3-dimensional independent  mean zero input signal with  covariance matrix $\Sigma _z$ given by  ${\rm vech}(\Sigma _z ) = (0.005,  0.0046, 0.0041, 0.0042, 0.0037, 0.004)$. In these figures the input gain $ \gamma = 0.3724$ and the phase shift $\phi = 0.7356$ are kept constant. The values of the input mask $C\in \mathbb{M} _{20, 3}$ are chosen randomly using a uniform distribution on the interval $\left[ -1, 1 \right] $. The left and middle panels show the error surfaces exhibited by the discrete and continuous time reservoirs computed via Monte Carlo simulations. The right panel shows the error produced by the reservoir model and computed evaluating  the explicit capacity formula that can be written down in that case.}
\label{multidimensional Ikeda}
\end{figure}

\noindent In the first empirical experiment we present a quadratic memory task to an individually operating TDR. More explicitly, we inject in the reservoir a three dimensional independent input signal $\{ {\bf z} (t)\}_{t \in \Bbb Z}$, ${\bf z} (t) \in \mathbb{R}^3 $  with  mean zero and  covariance matrix $\Sigma_z$, and we study its ability to reconstruct the signal $y(t) = \sum_{h=0}^3 \sum_{i ,j=1}^3 z_i(t-h) z_j(t-h) \in \mathbb{R} $. In the terminology introduced in Section \ref{Multidimensional memory tasks}, this exercise amounts to a quadratic task characterized by the vector $Q \in \mathbb{R} ^{78}$ defined by  $Q:= ({\rm vec}(Q ^{\ast} ))^{\top} D_{12} $, with $D_{12} $ the duplication matrix in dimension twelve and $Q ^\ast\in \Bbb S _{12}$ a block diagonal matrix with four  matrices $\mathbf{i} _3 \mathbf{i} _3 ^\top$ as its diagonal blocks. Indeed, if ${\bf z} ^3(t):= \left({\bf z} (t), {\bf z} (t-1), {\bf z}(t-2), {\bf z}(t-3)\right)  $, it is clear that
\begin{eqnarray*}
y(t)&=&  {\bf z} ^3(t)  ^{\top}Q ^\ast {\bf z} ^3(t) = {\rm trace} \left( {\bf z} ^3(t)  ^{\top}Q ^\ast {\bf z} ^3(t)\right)= ({\rm vec}(Q ^{\ast\, \top} ))^{\top} \left({\bf z} ^3(t)\otimes \mathbb{I}_{12}\right){\rm vec} \left({\bf z} ^3(t)\right)\\
&=&  ({\rm vec}(Q ^{\ast} ))^{\top} {\rm vec} \left({\bf z} ^3(t){\bf z} ^3(t) ^{\top}\right)
=  ({\rm vec}(Q ^{\ast} ))^{\top} D_{12}{\rm vech} \left({\bf z} ^3(t){\bf z} ^3(t) ^{\top}\right)=Q \cdot {\rm vech} \left({\bf z} ^3(t){\bf z} ^3(t) ^{\top}\right).
\end{eqnarray*}
In order to tackle this multidimensional task we use two twenty neuron TDRs constructed using the Mackey-Glass \eqref{Mackey-Glass nonlinear kernel} and the Ikeda \eqref{Ikeda nonlinear kernel} nonlinear kernels.
In Figures \ref{multidimensional MG} and \ref{multidimensional Ikeda} we depict the error surfaces exhibited by both RCs in discrete and continuous time as a function  of the distance between neurons and the feedback gain $\eta $ and using fixed input gains $\gamma $ whose values are indicated in the legends. Those error surfaces are computed using Monte Carlo simulations. At the same time we compute the error surfaces produced by the corresponding reservoir model and by evaluating  the explicit capacity formula that can be written down in that case. The resulting figures exhibit a remarkable similarity that had already been observed for scalar input signals in~\cite{GHLO2014_capacity}. More importantly, the figures show the ability of the theoretical formula based on the reservoir model to locate the regions in parameter space for which the reservoir performance is optimal.

\medskip
   
\noindent {\bf {Robustness properties of the parallel reservoir architecture.}} 

\medskip

\noindent We now study using the parallel reservoir model introduced in Section~\ref{The approximating model of parallel TDRs with multidimensional input signals} the robustness properties of the parallel architecture with respect to parameter choice and misspecification task. 

\medskip

\noindent {\bf (i) Parallel TDR configurations and robustness with respect to the choice of reservoir parameters.} An interesting feature of parallel TDR architectures that was observed in \cite{GHLO2012} is that optimal performance has a reduced sensitivity with respect to the choice of reservoir parameters when compared to that of individually operating reservoirs. In order to provide additional evidence of this fact, we have constructed parallel pools of  2, 5, 10, and 20 parallel Mackey-Glass-based TDRs and we present to them a nine-lag quadratic memory task is $y(t) = \sum^{9}_{i = 0} z(t-i)^2$ that, in the terminology introduced in Section \ref{Multidimensional memory tasks} corresponds to the quadratic task characterized by the vector $Q= \left({\rm vec}(\mathbb{I}_{10})\right)^{\top} D_{10}$, with $D_{10}$ the duplication matrix in dimension ten. For each of these parallel TDR architectures, as well as for an individually operating TDR, we will vary the number of the constituting neurons from 20 to 100. For each of these resulting configurations  we randomly draw 1000 sets of input masks with entries uniformly distributed in the interval  $[-3,3]$ and reservoir parameters and distance between neurons $d$, also uniformly distributed in the intervals $\eta\in \left[ 1,3 \right] $, $\gamma \in [-3,3]$, and $d\in (0, 1)$. 

\begin{figure}[h]
\hspace{-2.4cm}\includegraphics[scale=.4, angle=0]{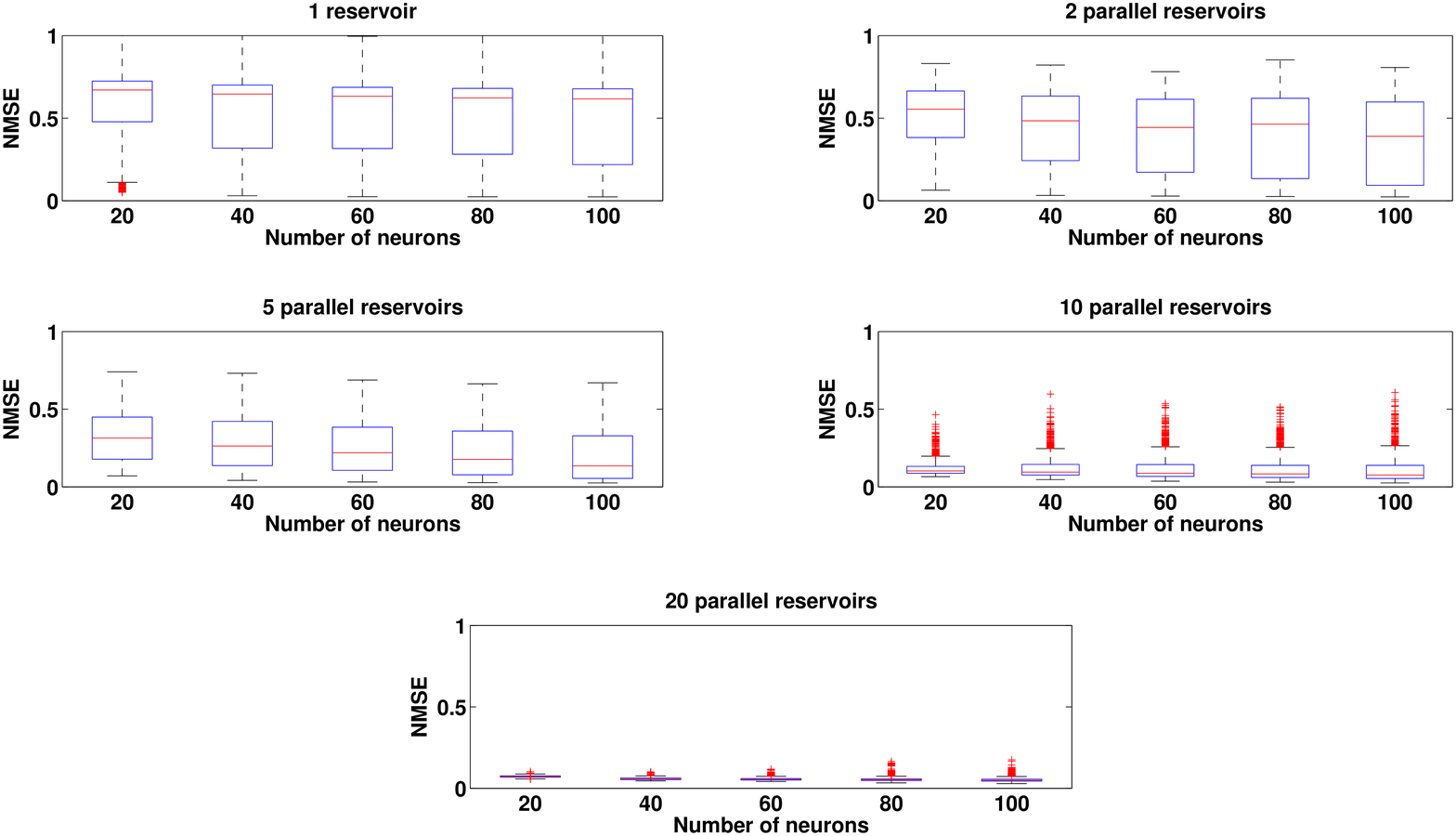}
\caption{Performance distributions of parallel arrays of Mackey-Glass-based  TDRs in a 9-lags quadratic memory task   when the kernel parameters and the input mask are varied randomly.}
\label{performance exercise misspecification}
\end{figure}
Figure \ref{performance exercise misspecification} provides the box plots corresponding to the distributions of normalized mean squared errors obtained with each configuration by making the input masks and reservoir parameter values  randomly vary, all of them computed using the capacity formulas associated to the parallel reservoir model \eqref{VAR model reservoir array}. This figure provides striking evidence of the facts that first, the parallel architecture performs on average better and second, that this performance is not sensitive to the choice of reservoir parameters and input mask.

\medskip

\begin{figure}[h]
\hspace{-1.5cm}\includegraphics[scale=.35, angle=0]{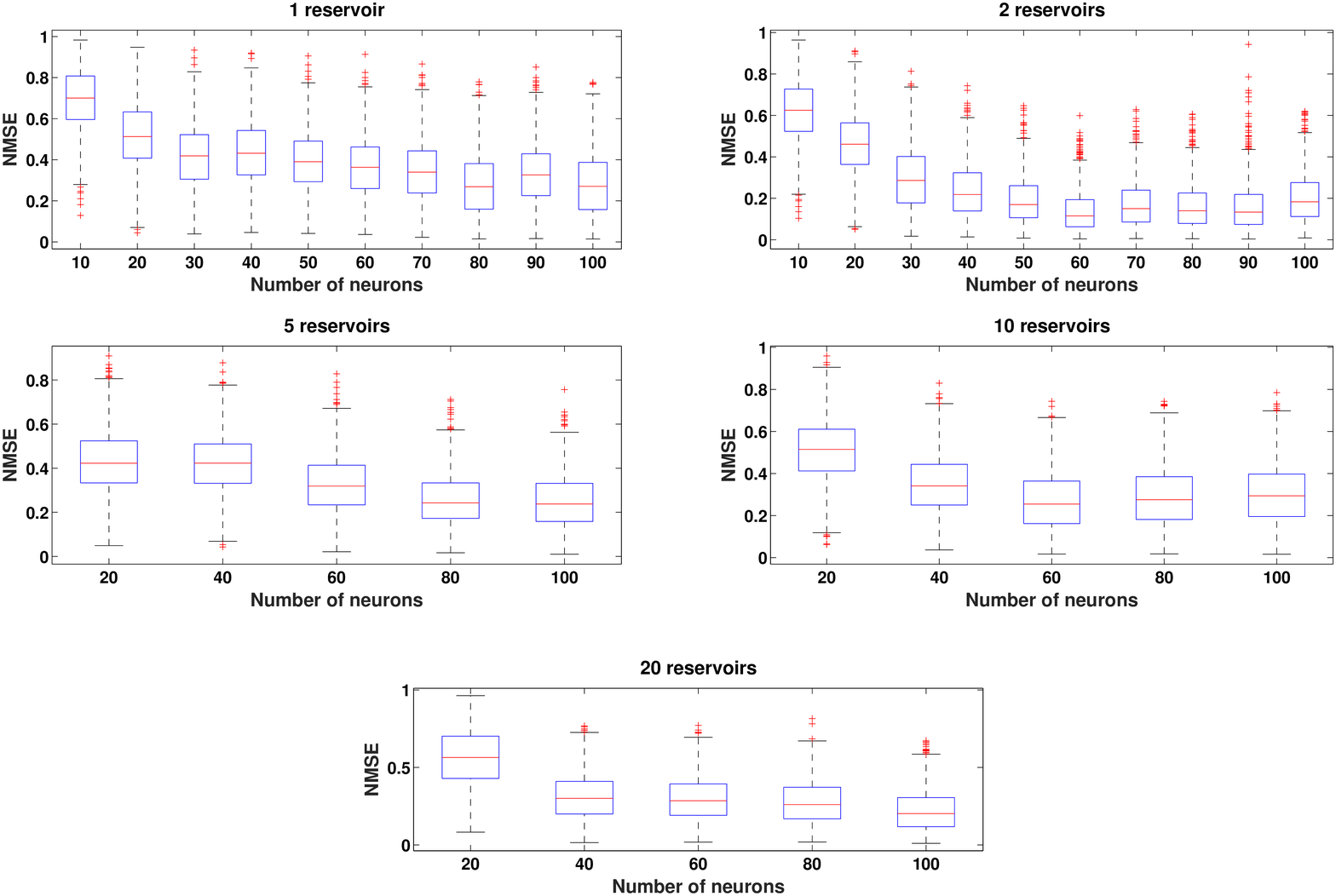}
\caption{Robustness of the performance of parallel arrays of Mackey-Glass kernel based TDRs with respect to  task misspecification. The box plots report the distribution of normalized mean square errors committed  in the execution of 1000 randomly generated 9-lag diagonal quadratic memory tasks by different TDR configurations that had been initially optimized for a 3-lag quadratic memory task.}
\label{performance exercise misspecification task}
\end{figure}

\noindent {\bf (ii) Parallel TDR configurations and robustness with respect to memory task misspecification.} The goal of the next experience  is to see how, given a specific memory task and given a parallel array of TDRs or an individually operating one that have been optimized for that particular task, the performance of the different configurations degrades when the task is modified but not the reservoir parameters. We use again parallel pools of 
1, 2, 5, 10, and 20 Mackey-Glass-based TDRs with neurons ranging from 10 to 100. For each of those configurations, we choose parameters that optimize their performance with respect to the 3-lag quadratic memory task specified by the matrix $Q= \left({\rm vec}(\mathbb{I}_{4})\right)^{\top} D_{4}$ and with respect to a one-dimensional independent input signal with mean zero and  variance $0.0001$.
Once the optimal parameters for each configuration have been found using the nonlinear capacity function based on the reservoir model \eqref{VAR model reservoir array}, we fix them we subsequently expose the corresponding reservoirs  to different randomly specified memory tasks of a different specification, namely, 1000 different 9-lag quadratic tasks of the form $Q= \left({\rm vec}(Q ^\ast _{10})\right)^{\top} D_{10}$, where $Q ^\ast \in \mathbb{M}_{10}$  is a randomly generated diagonal matrix with entries drawn from the uniform distribution  on the interval $[-10;10]$. Figure \ref{performance exercise misspecification task} contains the box plots corresponding to the performance distributions of the different configurations. In this case, parallel architectures offer an improvement on performance and robustness when compared to the single reservoir design. The most visible improvement is obtained in this case when using a parallel pool of two reservoirs.

\end{document}